\documentclass[journal,twoside,web]{ieeecolor}

\usepackage{etoolbox}
\makeatletter
\@ifundefined{color@begingroup}%
{\let\color@begingroup\relax
\let\color@endgroup\relax}{}%
\def\fix@ieeecolor@hbox#1{%
\hbox{\color@begingroup#1\color@endgroup}}
\patchcmd\@makecaption{\hbox}{\fix@ieeecolor@hbox}{}{\FAILED}
\patchcmd\@makecaption{\hbox}{\fix@ieeecolor@hbox}{}{\FAILED}
\usepackage{tmi}
\usepackage{cite}
\usepackage{amsmath,amssymb,amsfonts}
\usepackage{algorithmic}
\usepackage{graphicx}
\usepackage{textcomp}
\usepackage{algorithm}
\usepackage{algorithmic}
\usepackage{multirow}
\usepackage{multicol}
\usepackage{utfsym}
\usepackage{array}
\usepackage{url}
\usepackage{booktabs}
\usepackage[table,xcdraw]{xcolor}   
\usepackage[hypertexnames=false]{hyperref}
\def\BibTeX{{\rm B\kern-.05em{\sc i\kern-.025em b}\kern-.08em
    T\kern-.1667em\lower.7ex\hbox{E}\kern-.125emX}}
\begin{document}

\bstctlcite{IEEEexample:BSTcontrol}
\title{CXRAgent: Director-Orchestrated Multi-Stage Reasoning for Chest  X-Ray Interpretation} 
\author{Jinhui Lou, Yan Yang, \IEEEmembership{Member, IEEE}, Zhou Yu, \IEEEmembership{Member, IEEE}, Zhenqi Fu, Weidong Han, Qingming Huang, \IEEEmembership{Fellow, IEEE}, and Jun Yu, \IEEEmembership{Senior Member, IEEE}
\thanks{This paper has been accepted for publication in IEEE Transactions on Medical Imaging. \copyright\ 2026 IEEE. Personal use of this material is permitted. Permission from IEEE must be obtained for all other uses, in any current or future media, including reprinting/republishing this material for advertising or promotional purposes, creating new collective works, for resale or redistribution to servers or lists, or reuse of any copyrighted component of this work in other works. }
\thanks{Jinhui Lou, Yan Yang, Zhou Yu  are with the School of Computer Science, Hangzhou Dianzi University, Hangzhou, 310018, China (e-mail: 22081034@hdu.edu.cn;  yangyan@hdu.edu.cn; yuz@hdu.edu.cn).} 
\thanks{Zhenqi Fu is with the Department of Automation, Tsinghua University, Beijing 100084, China (e-mail: fuzhenqi@mail.tsinghua.edu.cn).}
\thanks{Weidong Han is with the Department of Colorectal Medical Oncology, Zhejiang Cancer Hospital, Hangzhou, 310022, China (e-mail: hanwd@zjcc.org.cn).} 
\thanks{Qingming Huang is with the School of Computer and Control Engineering, University of Chinese Academy of Sciences, Beijing, 101408, China (e-mail: qmhuang@ucas.ac.cn). }
\thanks{Jun Yu is with the School of Intelligence Science and Engineering, Harbin Institute of Technology (Shenzhen), 518055, China. (e-mail: yujun@hit.edu.cn).  (Corresponding authors: Yan Yang and Jun Yu) }
} 
\maketitle

\begin{abstract}
Chest X-ray (CXR) plays a pivotal role in clinical diagnosis, and a variety of task-specific and foundation models have been developed for CXR interpretation. However, these models often struggle to adapt to new diagnostic tasks and complex reasoning scenarios. Recently, LLM-based agents have emerged as a promising paradigm for CXR analysis, enhancing model's capability via tool coordination, multi-step reasoning, and team collaboration, etc. However, existing agents often rely on a single diagnostic pipeline and lack mechanisms for assessing tools' reliability, limiting their adaptability and credibility. To this end, we propose CXRAgent, a director-orchestrated, multi-stage agent for CXR interpretation, where a central director coordinates the following stages: (1) Tool Invocation: The agent strategically orchestrates a set of CXR-analysis tools, with outputs normalized and verified by the Evidence-driven Validator (EDV), grounding diagnostic outputs with visual evidence to support reliable downstream diagnosis; (2) Diagnostic Planning: Guided by task requirements and intermediate findings, the agent formulates a targeted diagnostic plan, assembles an expert team, defines member roles, and coordinates their interactions to enable adaptive collaborative reasoning; (3) Collaborative Decision-making: The agent integrates insights from the expert team with accumulated contextual memories, synthesizing them into an evidence-backed  conclusion. \textcolor{black}{Experiments on diverse tasks show that CXRAgent achieves strong performance with reliable visual grounding, attaining overall accuracies of 67.0\% on CheXbench and 75.6\% on Medical-CXR-VQA, and a RaTEScore of 0.569 on MIMIC-CXR for report generation.}  Code and data are available at this \href{https://github.com/laojiahuo2003/CXRAgent/}{link}.
\end{abstract} 

\begin{IEEEkeywords}
Chest X-ray interpretation, medical agent, large multimodal model, multi-stage reasoning.
\end{IEEEkeywords}

\section{Introduction}
\label{sec:introduction}

\IEEEPARstart{C}{hest} X-ray (CXR) is among the most widely used imaging modalities in clinical practice due to its affordability, rapid acquisition, and diagnostic utility across a wide range of thoracic conditions. However, accurate interpretation demands substantial clinical expertise and years of specialized training. In addition, the heavy workload of radiologists makes the diagnostic process susceptible to delays and errors, including missed or incorrect findings. These challenges have motivated a growing interest in AI-powered systems that aim to assist clinicians and improve the efficiency and precision of CXR interpretation.

\begin{figure}[t]
\centering
\includegraphics[width=1\columnwidth]{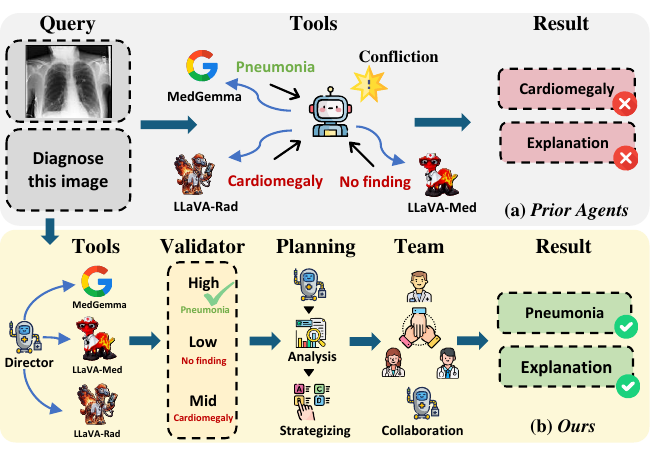}
\caption{Comparison between prior agent models and our CXRAgent. (a) Prior agents often struggle to deal with tool conflicts in complex cases and rely on a single diagnosis pipeline. (b) CXRAgent adopts a multi-stage pipeline, guided by a core director, to flexibly coordinate tool invocation,  tool-output validation, diagnostic planning, and team-integrated  collaborative decision-making. Each tool's confidence is validated with visual evidence to ensure reliability.}
\label{fig1}
\end{figure}

Previous efforts have primarily focused on developing task-specific models for automated CXR interpretation. Representative efforts in CXR report generation include RGRG \cite{24},  R2GenGPT \cite{WANG2023}, MAIRA-2 \cite{bannur2024maira}, and LLaVA-Rad \cite{zambrano2025clinically}. RGRG \cite{24} proposed to first detect anatomical structures and then integrate the region-level descriptions into a unified radiology report.  R2GenGPT \cite{WANG2023} introduced a lightweight fine-tuning strategy to achieve efficient vision-language alignment for radiology report generation. MAIRA-2 \cite{bannur2024maira} advanced grounded radiology report generation by incorporating other reporting contexts as additional inputs.  LLaVA-Rad  \cite{zambrano2025clinically}  trained a domain-adapted chest X-ray encoder and connected it to pre-trained language models through a lightweight adapter for vision-language alignment. While these task-specific models have achieved notable success in individual tasks, they struggle to generalize across different tasks (for example, from report generation to question answering). 

Recent progress in deep learning has further fostered the development of medical foundation models for CXR analysis. Examples include RadFM \cite{wu2023generalistfoundationmodelradiology}, CheXagent \cite{chen2024chexagent}, Ark\textsuperscript{+} \cite{ark}, and MedGemma \cite{medgemma-hf}. RadFM \cite{wu2023generalistfoundationmodelradiology} introduced a generalist foundation model capable of modality recognition, disease diagnosis, visual question answering, etc. CheXagent \cite{chen2024chexagent} curated a large-scale CheXinstruct dataset and developed a vision-language foundation model for CXR interpretation. Ark\textsuperscript{+} \cite{ark} employed a cyclical pretraining strategy to accumulate and reuse knowledge embedded in expert labels for CXR analysis. MedGemma-4B \cite{medgemma-hf} is a large multimodal model designed for medical applications and aims to serve as a tool for agent-based diagnostic systems. However, although these foundation models can perform well on predefined diagnostic tasks, they still struggle to generalize to new diagnostic objectives and complex reasoning scenarios.

Most recently, agent-based medical diagnostic models have gained increasing attention for their ability to unify the strengths of diverse models (e.g., task-specific models and foundation models) and mimic clinical reasoning workflows. These agents aim to advance diagnostic intelligence by enabling multi-step, context-aware reasoning and integrating diverse diagnostic tools. In particular, agent models decompose complex diagnostic tasks into a series of subtasks. By orchestrating tools step by step and solving individual subtasks, agents enhance the transparency of the diagnostic pipeline. Representative agent models in medical image analysis such as MedRAX \cite{fallahpour2025medrax}, MDAgents \cite{kim2024mdagents} and MMedAgent \cite{li2024mmedagent} have demonstrated the potential of agent-based systems by intelligently coordinating multiple analysis tools. However, they typically treat tools as equivalent components, lacking effective mechanisms to assess the reliability and expertise of each tool's output. As shown at the top of Fig.~\ref{fig1}, they struggle to reconcile the conflicting outputs across tools, thereby failing to prioritize more trustworthy information. This risks diluting critical diagnostic signals and undermines the diagnosis credibility. Moreover, many agents rely on a rigid single-stage workflow, lacking flexibility. 

\textcolor{black}{Regarding these limitations, our work is motivated by three core questions: 1) How can we build a unified agent framework that generalizes across diverse CXR interpretation tasks? 2) How can the agent assess and reconcile conflicting tool outputs through explicit visual evidence validation? 3) How can dynamic, team-based reasoning, inspired by clinical multidisciplinary teams (MDT), be integrated into an AI system to enhance adaptability and decision-making in complex cases? To answer these questions, as shown at the bottom of Fig.~\ref{fig1}, we introduce CXRAgent, a multi-stage, director-orchestrated agent tailored for evidence-backed and adaptive CXR interpretation.} Firstly, the agent orchestrates a set of CXR analysis tools, with their outputs reformatted and validated by an Evidence-driven Validator (EDV). The EDV evaluates the confidence of each tool’s output by examining supporting or refuting visual evidence, ensuring reliable inputs for downstream diagnosis. Then, the agent performs flexible planning by assembling an expert team and enabling adaptive collaborative reasoning tailored to  task requirements, thereby improving decision-making in complex clinical scenarios. The main contributions of our paper are as follows: 
\begin{itemize}

\item We propose CXRAgent, a multi-stage, director-orchestrated agent for chest X-ray interpretation that employs a powerful multi-modal LLM as its core director to flexibly coordinate tool invocation, diagnostic planning, and team-based collaborative decision-making.

\item  We propose the Evidence-driven Validator (EDV) that grounds tool results with visual supportive and refuting evidence, provides confidence assessments to evaluate diagnostic reliability and unifies the output formats of various tools to support reliable downstream diagnosis. 

\item Inspired by clinical Multidisciplinary Team (MDT) practices, we propose a flexible team-based collaborative paradigm that intelligently assembles and coordinates specialized agent teams, ensuring robust adaptability to diverse tasks and diagnostic complexities. 

\item Experimental results demonstrate that CXRAgent achieves substantial performance gains across various CXR interpretation tasks including visual question answering and report generation, validating its diagnostic accuracy and adaptability.  
\end{itemize}

\section{Related Works}

In this section, we review related works on chest X-ray analysis models and medical agent systems.

\subsection{Chest X-ray Analysis Models}
Recent studies in chest X-ray analysis have introduced numerous task-specific models aimed at distinct clinical objectives, such as classification \cite{cohen2022torchxrayvision}, detection \cite{Detection}, localization \cite{PIXEL}, segmentation \cite{ma2024segment}, visual question answering \cite{MVQA, FAVP}, and report generation \cite{Tokenmixer, WANG2023, bannur2024maira, zambrano2025clinically}.  PIXEL \cite{PIXEL} introduced a prompt-driven constrained generative framework that synthesizes anatomically aligned healthy–diseased image pairs to enable supervised learning of pathology localization. Building on visual prompting, FAVP \cite{FAVP} proposed an Adaptive Visual Prompt Creator that dynamically generates eight types of region-level visual prompts to improve CXR visual question answering. META-CXR \cite{11153468} introduced an abnormality-guided vision-language model using multimodal expert tokens,  extracting visual features from different encoders for  abnormality classification and report generation. MobileNetV2-based CXR classification \cite{9682248} utilized fine-tuned lightweight architectures for COVID-19 and pneumonia identification, yet it primarily captures global patterns without providing precise spatial localization. Transformer-based report generation \cite{10145699} integrated image segmentation to enhance feature extraction, but the single-stage structure may limit its capacity to reason over complex, multi-focal findings across anatomical regions. CvT2DistilGPT2 \cite{cvt2distilgpt2} proposed to leverage warm starting for report generation and IFCC \cite{ifcc} introduced two rewards to encourage factually complete and consistent  reporting. 
 
The advent of multimodal foundation models has further advanced CXR understanding and interpretation.
LLaVA-Med \cite{li2023llava} constructed a curated biomedical image–text instruction dataset and optimized a vision–language model tailored for biomedical tasks. \textcolor{black}{MAIRA-2 \cite{bannur2024maira} advanced grounded radiology report generation by incorporating other reporting contexts as additional inputs.}  LLaVA-Rad \cite{zambrano2025clinically} adopted a three-stage training paradigm with domain-specific encoder pre-training for efficient and clinically deployable CXR report generation. \textcolor{black}{CXRMate \cite{NICOLSON2024101585}  proposed a longitudinal, multi-image CXR report generator trained with the CXR-BERT reward. CXRMate-ED \cite{CXRMate-ED} investigated the use of patient data from emergency department records for report generation.  MedVersa \cite{MedVersa} proposed a generalist foundation model for CXR interpretation.  Libra \cite{libra} proposed a temporal-aware Multimodal LLM for CXR report generation.  Janus-Pro \cite{Janus-Pro} proposed a novel autoregressive framework that unifies multimodal understanding and generation. } 
MedFILIP \cite{liang2025medfilip} achieved fine-grained cross-modal alignment through entity extraction and knowledge injection, while Med-Flamingo \cite{moor2023med} leveraged high-quality interleaved data and few-shot learning to enhance adaptability. CheXagent \cite{chen2024chexagent} curated the large-scale CheXinstruct dataset and developed a vision–language foundation model specialized for CXR interpretation. Lingshu \cite{Lingshu} presented a generalist foundation model for unified multimodal medical understanding and reasoning.
DeepMedix-R1 \cite{DeepMedix-R1} introduced a holistic medical foundation model for CXR interpretation, following a sequential training strategy: it is first fine-tuned on curated instruction data to acquire fundamental interpretive skills, then trained on synthetic reasoning samples for cold-start reasoning, and finally refined via online reinforcement learning to enhance reasoning precision and generative quality. MedGemma-4B \cite{medgemma-hf} proposed a multimodal model designed for medical image understanding and reasoning, integrating visual and textual modalities to support diverse clinical tasks. 

Although these models have achieved impressive results across benchmarks, they encounter challenges in handling complex, ambiguous, or multi-condition cases. Their architectures often lack the flexibility required to dynamically adapt to diverse diagnostic tasks. However, these advances lay the groundwork for agent-based CXR analysis, which aspires to achieve  adaptive, interpretable, and complex reasoning. 

\subsection{Medical Agent Systems}
Medical agent systems \cite{jin2024agentmd,COD,huang2025biomni} have emerged as a promising paradigm to overcome the limitations of monolithic models by orchestrating multiple specialized components. MedAgents \cite{tang2024medagents} introduced a multi-disciplinary collaboration framework, where LLM-based agents engage in role-playing and multi-round discussions to enhance reasoning in medical tasks. MDAgents \cite{kim2024mdagents} proposed a multi-agent framework that automatically assigns task-specific solo or group collaboration structures to LLMs, emulating real-world medical decision-making processes. ClinicalAgent \cite{yue2024clinicalagent} applied this agentic paradigm to clinical trial prediction, while DoctorAgent-RL \cite{DoctorAgent} modeled clinical consultations as dynamic decision-making under uncertainty using multi-agent reinforcement learning. Deep-DxSearch \cite{zheng2025end} proposed end-to-end reinforcement learning to jointly optimize retrieval and reasoning, enabling retrieval-aware diagnostic strategies beyond inference-only agentic systems. \textcolor{black}{DeepRare~\cite{Zhao2026DeepRare} presented a multi-agent framework for rare disease diagnosis with traceable reasoning. It integrated multiple specialized tools and up-to-date medical knowledge sources to process heterogeneous clinical inputs, including free-text clinical notes, structured Human Phenotype Ontology terms, and genetic testing results. The system generated ranked differential diagnostic hypotheses with explicit reasoning chains grounded in verifiable medical evidence.}

Recently, multi-modal agents for medical image analysis have also been developed. MMedAgent \cite{li2024mmedagent} proposed to manage diverse medical tasks by integrating open-source medical models. It consists of an instruction-tuned multi-modal LLM serving as an action planner and results aggregator, and a set of task-specific medical tools. \textcolor{black}{While this design enables multi-task and multi-modal capabilities, the system relies on largely predefined workflows and lacks explicit mechanisms to resolve conflicting outputs or ensure traceable, evidence-grounded reasoning. }MMA \cite{peng2024integration} conducted medical diagnosis question answering by deploying specialized LLM-based agents that processed multi-source data and collaborated via structured workflows. MedRAX \cite{fallahpour2025medrax} demonstrated the potential of agent-based systems by orchestrating multiple diagnostic tools for CXR interpretation. \textcolor{black}{However, due to the absence of an explicit validation mechanism to verify intermediate findings, the integration of multiple tools may introduce inconsistent outputs, which can lead to performance degradation. }MAM \cite{zhou2025mam} proposed a modular multi-agent architecture for multimodal medical diagnosis, leveraging role decomposition and collaborative decision-making. \textcolor{black}{But interactions among agents largely follow  predefined workflows and lack mechanisms for resolving conflicting conclusions. } CT-Agent \cite{mao2025ct} incorporated anatomy-aware reasoning with hierarchical token compression to efficiently handle 3D CT analysis. PathoAgenticRAG \cite{zhang2025patho} presented a multimodal retrieval-augmented generation framework for pathology vision-language models, mitigating hallucinations via a database of page-level embeddings from pathology textbooks. Agentic paradigms have also been extended to other medical applications. For example,  Xie \textit{et al}. \cite{xie2025prompt} proposed a decentralized multi-agent reinforcement learning strategy to interact dynamically with a foundation model, mitigating its undue influence and progressively refining priors for low-count PET reconstruction. 

Despite these advances, current medical imaging diagnosis agent systems face persistent challenges: (1) insufficient mechanisms to reconcile conflicting outputs across tools, (2) reliance on a single, rigid workflow that ignore case complexity, and (3) weak or absent connections between diagnostic conclusions and supporting visual evidence.
\begin{figure*}[t]
\centering
\includegraphics[width=1\textwidth]{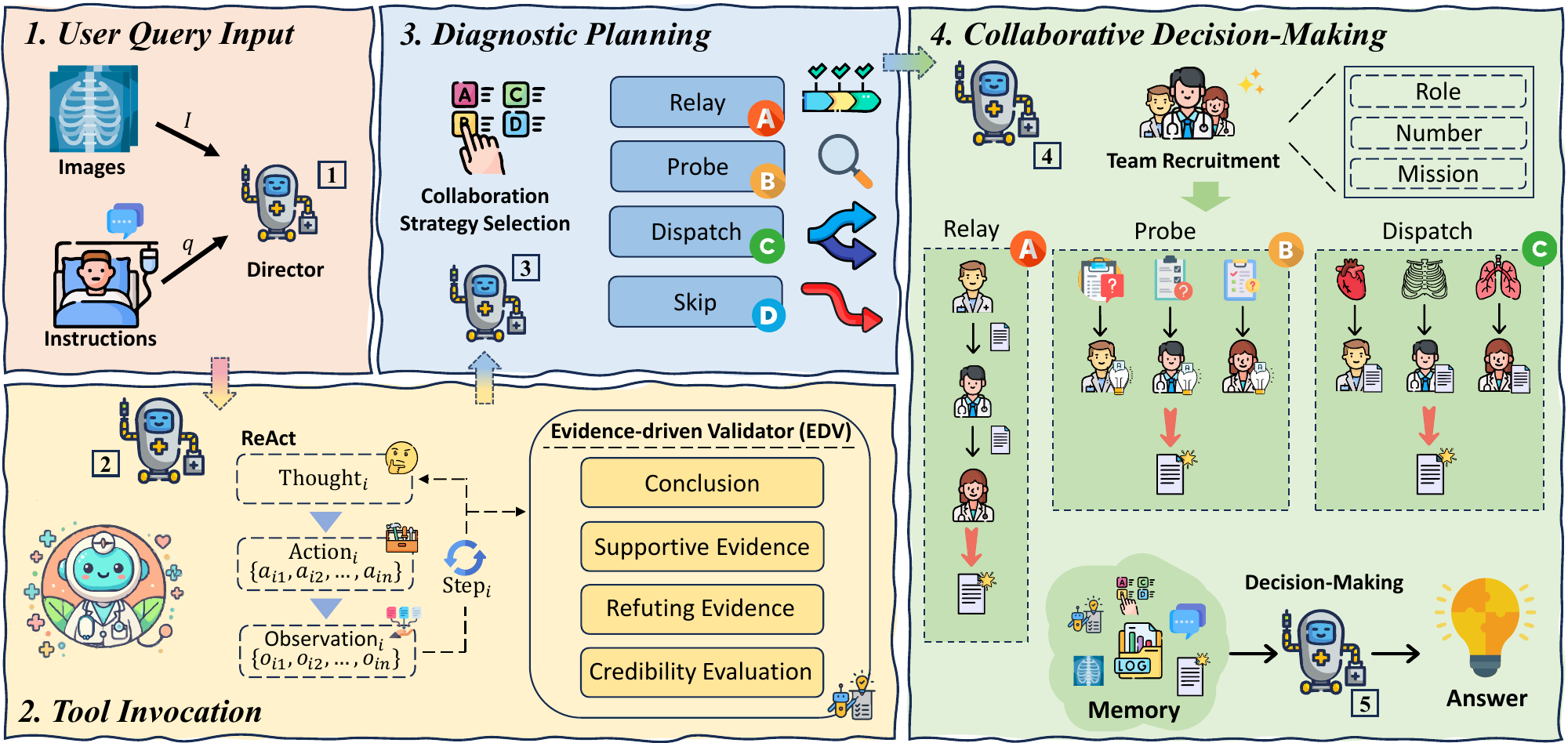}

\caption{\textbf{Overview of the CXRAgent}. The central director coordinates the  following  key stages: (1) ingesting user queries, (2) orchestrating CXR-analysis tools with outputs normalized and validated by the Evidence-driven Validator (EDV) to ensure visually grounded reliability, (3) making diagnostic plans to adaptively assemble specialized expert teams for collaborative reasoning, and (4) synthesizing evidence-backed conclusions by integrating expert collaborative insights and the contextual memories.}
\label{fig2}
\end{figure*}
\section{Method}

As illustrated in  Fig.~\ref{fig2}, CXRAgent is designed as a multi-stage, director-orchestrated agent for chest X-ray interpretation. Upon receiving a user query, it leverages a multi-modal LLM as the central ``director" to coordinate the following key stages: 
(1) \textbf{Tool Invocation} — The agent orchestrates multiple CXR-analysis tools through iterative reasoning cycles. Multimodal inputs are analyzed step by step, and the outputs of various tools are normalized and validated by the Evidence-driven Validator, which aligns tool responses with corresponding visual evidence to provide reliable insights for downstream reasoning.
(2) \textbf{Diagnostic Planning} — Guided by task requirements and intermediate results, the agent formulates subsequent diagnostic plans that adapt to evolving clinical needs and varying case complexities.
(3) \textbf{Collaborative Decision-Making} — Guided by the diagnostic plan and task objectives, the agent assembles a team of specialized experts to execute the required analyses, integrating insights from team decisions and historical reasoning trajectories to deliver a comprehensive, evidence-grounded diagnostic conclusion.

\subsection{Tool Invocation}
We first employ the ReAct \cite{yao2023react} to iteratively plan and execute tool invocations, followed by the Evidence-driven Validator (EDV) to refine and evaluate the tool outputs with visual evidence. \textcolor{black}{In our framework, the tool selection  is not a one-time predefined sequence but rather a dynamic and reflective cycle that evolves based on the specific context of each case. The Director begins by performing a deep semantic analysis of the user prompt to identify the core clinical objective. This initial selection is driven by the complexity of the query. Tool invocation is then iterated and can be terminated dynamically based on real-time feedback stored in the context memory.}  Concretely, ReAct contains the following steps: 

\textbf{Thought Step – State Analysis.}
At each iteration, the agent decides whether additional tools are needed. The decision is jointly conditioned on the input images $I$, the textual query $q$, and the accumulated reasoning memory log, denoted as $\text{Log}_{i-1}$:

\begin{equation}
\text{Thought}_i = \text{Plan}(\text{Log}_{i-1}, I, q).
\end{equation}

\noindent If the agent determines that invoking additional tools is necessary to gather sufficient information under the current iteration ${i}$ for diagnosis, it proceeds to the subsequent stages.

\textbf{Action Step – Tool Invocation.}
The model selects the most appropriate tool(s) based on the current reasoning context, formulated as:

\begin{align}
\text{Action}_i &= \{a_{i1}, a_{i2}, \ldots, a_{in}\} \\
&= \bigcup_{j=1}^{n} \text{Tool}_{ij}(I, q, \text{Thought}_i), \quad \text{Tool}_{ij} \in \mathcal{T}
\end{align}
where $\mathcal{T}$ is the tool set, $\text{Tool}_{ij}$ is the $j$-th tool of $i$-th iteration. $ a_{i1}, a_{i2}, \ldots, a_{in}$ are the selected tools (i.e., arrangement of tool invocation), $n$ is the total number of tools invoked. All invocations are logged for further reasoning. 

\textbf{Observation Step – Result Aggregation.}
The selected tools $\{a_{i1}, a_{i2}, \ldots, a_{in}\}$ are applied to the relevant image-query pairs, producing results $\{o_{i1}, o_{i2}, \ldots, o_{in}\}$, which are then aggregated as observations to update the reasoning log for the next iteration. These observations will provide new information for the next iteration of the reasoning process:

\begin{equation}
\text{Obs}_i = \{o_{i1}, o_{i2}, \ldots, o_{in}\},
\end{equation}
\begin{equation}
\text{Log}_i \leftarrow \text{Log}_{i-1} \cup \text{Obs}_i.
\end{equation}
\textbf{Thought Step – Reflection and State Analysis.}
After gathering the results, the model reflects on the new observations and updates its reasoning state. This process determines whether further tool invocations are necessary:

\begin{equation}
\text{Thought}_{i+1} = \text{Plan}(\text{Log}_{i}, I, q).
\end{equation}

\noindent{The model repeatedly performs the thought, action, and observation steps until sufficient information is accumulated.}

\textbf{Evidence-driven Validator (EDV).} To mitigate diagnostic errors arising from unreliable tool outputs, we introduce the Evidence-driven Validator, a tool-agnostic verification module. Unlike methods that merely summarize tool results, EDV actively grounds results with visual evidence from the input image. For each diagnostic statement produced by a tool, EDV identifies supporting and contradicting evidence from the image and estimates the overall confidence of the statement. It also standardizes the statements of tools. After validating and normalizing outputs from individual tools, EDV enhances overall reliability of the diagnostic process. Concretely, EDV generates and normalizes a set of structured components that explain, validate, and assess the credibility of each diagnostic result:

\begin{itemize}
    \item \textbf{Conclusion}: A concise reformatted restatement of the diagnostic statement produced by the tool.

    \item \textbf{Supportive Evidence}: \textcolor{black}{Key image findings that  confirm and reinforce the diagnostic statement. (e.g., the presence of blunting of the costophrenic angle or fluid in the pleura would support a finding of ``Pleural Effusion").}
    \item \textbf{Refuting Evidence}: Observations that contradict the statement, including absent key evidence or contradictory cues. \textcolor{black}{(e.g., the absence of an increased cardiothoracic ratio would refute ``Cardiomegaly'').}
    \item \textbf{Confidence Assessment}: A qualitative judgment of the statement’s reliability, based on both the supportive and refuting evidence.
\end{itemize}

The structured and reformatted outputs produced by EDV establish a unified framework for validating and explaining predictions across different diagnostic tools. By explicitly grounding each conclusion in the visual evidence from the input data, EDV not only improves transparency but also enables consistent, interpretable cross-tool evaluation for the downstream interpretation. \textcolor{black}{We clarify that EDV does not rely on a secondary, task-specific vision model. Instead, it is implemented as a high-level reasoning protocol executed by the central multi-modal LLM (i.e., the Director), which semantically parses the tool’s output and performs cross-modal verification using its visual and medical knowledge. In our framework, the visual evidence is defined as localized radiographic features in accordance to the finding reported by the tool, categorized as the above Supportive Evidence and Refuting Evidence. The identified visual evidence satisfies the following criteria: The evidence must be visible within the radiographic image, rather than inferred from prior context. The localized image features must correspond unambiguously to the proposed abnormality, ensuring that the claim is grounded in radiographic signs. There should be no clear contradictory visual information in the image that would conflict with the asserted abnormality. The evidence must provide sufficient information to justify downstream diagnostic decisions, enabling the system to confidently incorporate the proposition into further diagnostic synthesis. If the image lacks radiographic signs that support the proposed conclusion, or if clear contradictory visual information is present, the result of the tool is assigned low confidence and is not directly used for downstream diagnostic synthesis.}

\textcolor{black}{Based on the visual cross-check, the EDV categorizes the confidence of each conclusion into three distinct levels: ``High", ``Medium" and ``Low". If the EDV assigns a ``Low'' confidence or identifies significant refuting evidence, the Director will initiate an iterative reasoning cycle, exploring alternative tools:}
\begin{itemize} 
    \item \textcolor{black}{\textbf{Dynamic Re-invocation:} The Director recognizes the potential failure of the current tool and adaptively calls an alternative tool with similar functionality but a different underlying architecture.}
    \item \textcolor{black}{\textbf{Conflict Resolution:} This multi-tool cross-check is the key to resolving conflicting conclusions. By comparing results from multiple sources, the Director can reason over agreements and discrepancies across tools, enabling an explicit assessment of inter-tool consistency.}
    \item \textcolor{black}{\textbf{Cautious Synthesis:} In cases where ambiguity remains, the Director is programmed to provide a ``cautious interpretation'' in the final report, explicitly noting the visual discrepancies to alert the radiologist, thereby mimicking the professional skepticism of a senior clinician.}
\end{itemize}

\subsection{Diagnostic Planning}

In real-world chest X-ray diagnostics, user queries vary widely in complexity and requirements, necessitating adaptable reasoning strategies and collaboration among multidisciplinary teams.  To this end, we design Diagnostic Planning, a stage guided by the ``director"  that dynamically determines collaboration modes (i.e., how to adaptively assemble specialized expert teams and formulate reasoning strategies) based on task-specific characteristics. We define four key collaboration modes:

\begin{enumerate}

\item \textbf{[Skip]} — Directly generates diagnostic results without team collaboration for those straightforward cases with unambiguous visual evidence, maximizing the diagnosis efficiency.

\item \textbf{[Relay]} — Employs a sequential refinement process where each specialist incrementally improves the diagnosis. Each expert builds on the outputs of the previous one, progressively refining conclusions. This mode is ideal for routine cases that require step-by-step refinement.

\item \textbf{[Dispatch]} — Decomposes complex tasks into specialized subtasks, assigning each to domain-specific experts for parallel analysis, enabling focused feature analysis. For example, assigning a heart expert for cardiac diagnosis and a bone expert for skeletal diagnosis.

\item \textbf{[Probe]} — Breaks down the case into targeted probing questions, prompting each team member to respond individually, facilitates the gathering of diverse and focused insights, supporting comprehensive understanding and diagnostic reasoning.
\end{enumerate}

For cases with clear diagnostic features, the system employs the Skip strategy to deliver immediate results without activating the team, maximizing efficiency for straightforward findings. When presented with well-defined diagnostic queries supported by unambiguous visual evidence, the system applies the Relay strategy, sequentially refining the diagnosis through stepwise expert analysis to ensure efficient and incremental improvement. For tasks requiring multi-faceted feature analysis, the Dispatch strategy engages specialized experts working in parallel, each focusing on distinct subtasks for targeted and efficient processing. Finally, for complex and ambiguous cases demanding high-level reasoning, the Probe strategy actively generates targeted probing diagnostic questions, enabling them to uncover subtle clues and collaboratively build a  well-supported diagnostic conclusion. \textcolor{black}{The strategy selection is not determined by a fixed memory table or a separately-trained policy model. Instead, the Director dynamically decides the strategy at inference time through task-aware, in-context reasoning and utilization of its context memory.}

\begin{figure}[h]
\centering
\includegraphics[width=1\columnwidth]{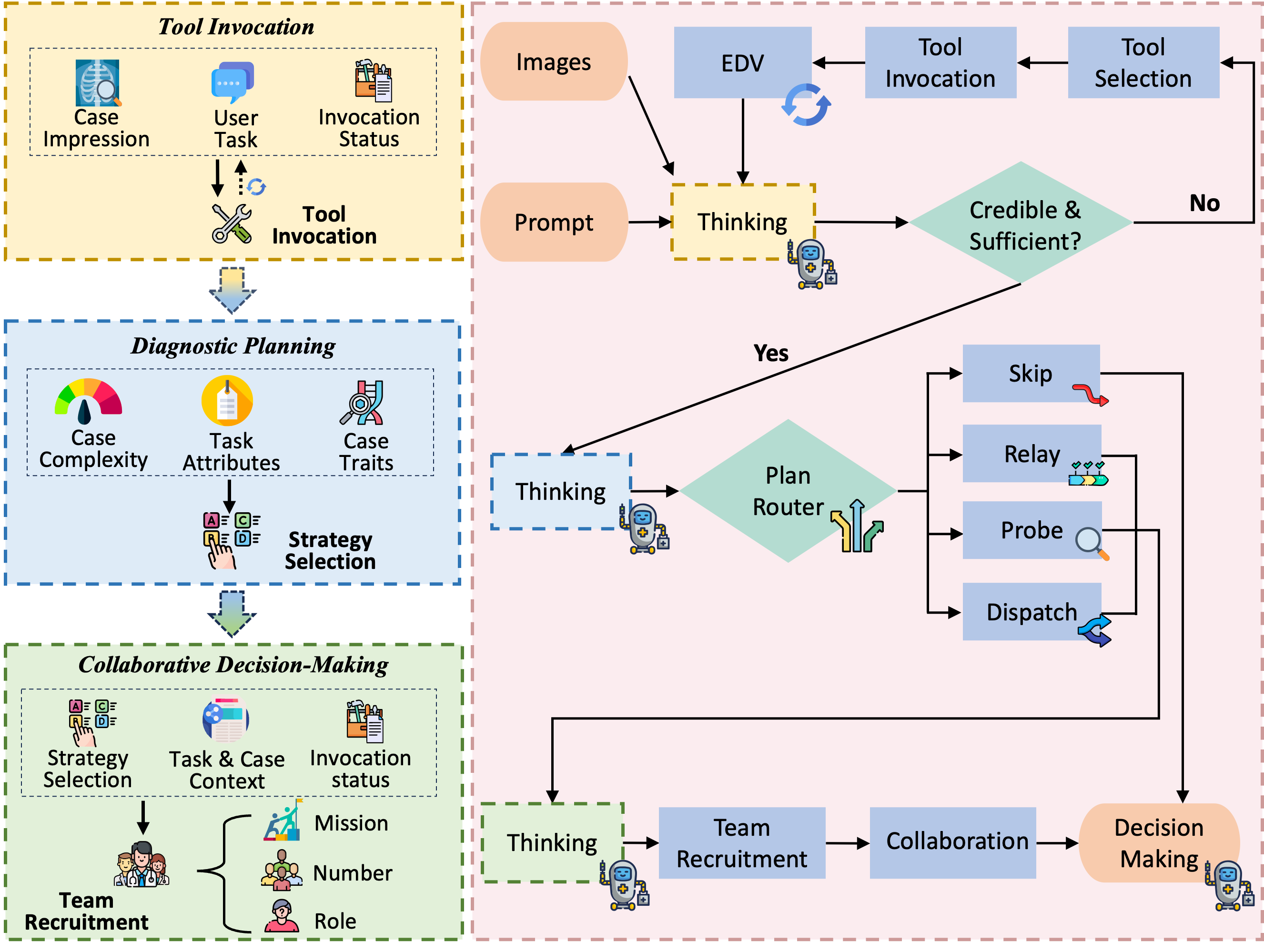}
\caption{Detailed flowchart of CXRAgent with iterations.}
\label{new}
\end{figure}
\subsection{Collaborative Decision-Making}
Based on the collaboration strategy determined by the Diagnostic Plan and the characteristics of the clinical case, the agent recruits a diagnostic team, intelligently configuring the number of team members, their roles, and missions. This configuration enables context-aware collaboration across diverse clinical scenarios. By aligning team composition with case complexity and diagnostic requirements, the system enables flexible workflows and delivers more reliable diagnostic results. Team recruitment can be formulated as: 

\begin{equation}
\text{Team} = \{ (\text{agent}_i, \text{role}_i, \text{mission}_i) \}_{i=1}^{k} = \mathcal{R}(I, q, s),
\end{equation}
where $\mathcal{R}$ denotes the team recruitment strategy, $s \in \{\text{Skip}, \text{Relay}, \text{Dispatch}, \text{Probe}\}$ is the selected collaboration strategy, and each tuple $\text{Agent}_i = (\text{agent}_i, \text{role}_i, \text{mission}_i)$ specifies the identity, functional role, and assigned mission of the $i$-th team member. Here, $k$ indicates the team size. This formulation allows dynamic configuration of diagnostic teams tailored to the complexity and specific demands of each clinical case. The variable $\text{mission}_i$ represents the sub-mission allocated to each agent, which varies by strategy: refinement tasks in \text{Relay}, dedicated subtasks in \text{Dispatch}, and probing questions in \text{Probe} mode. 

Algorithm~\ref{alg:amtb} demonstrates the complete workflow of team formation and collaborative diagnosis. After collecting EDV-validated tool outputs, team collaboration results, and the contextual memories, CXRAgent makes the final diagnostic decision through evidence-based reasoning. \textcolor{black}{Detailed flowchart of CXRAgent is shown in Fig.~\ref{new}. The framework first iteratively selects and validates diagnostic tools, then plans diagnostic strategies, and orchestrates team-based agent collaboration to generate evidence-backed radiographic interpretations.}
\begin{algorithm}[t]
\caption{Collaborative Decision-Making}
\label{alg:amtb}
\textbf{Input}: Query $q$, CXR images $I$, Strategy $s$, Memory $M$\\
\textbf{Output}: Diagnosis Result
\begin{algorithmic}[1]
\STATE $ \text{Team} \gets \mathcal{R}(q, I, s)$
\IF{$s = \text{Skip}$}
    \STATE $\text{TeamOutput} \gets \text{None}$ 
\ELSIF{$s = \text{Relay}$}
    \STATE $r_0 \gets \emptyset$
    \FOR{$i = 1$ to $k$}
        \STATE $r_i \gets \text{Agent}_i(I, r_{i-1}, \text{mission}_i := \text{task}_i)$
    \ENDFOR
    \STATE $\text{TeamOutput} \gets r_k$
\ELSIF{$s = \text{Dispatch}$}
    \FOR{$i = 1$ to $k$}
\STATE $r_i \gets \text{Agent}_i(I, \text{mission}_i := \text{sub-task}_i)$ 

    \ENDFOR
    \STATE $\text{TeamOutput} \gets \text{Concat}(\{r_1, r_2, \dots, r_k\})$
\ELSIF{$s = \text{Probe}$}
    \FOR{$i = 1$ to $k$}
        \STATE $r_i \gets \text{Agent}_i(I, \text{mission}_i := \text{question}_i)$ 
    \ENDFOR
    \STATE $\text{TeamOutput} \gets \text{Concat}(\{r_1, r_2, \dots, r_k\})$   
\ENDIF
\STATE $ \text{Diagnosis Result} \gets \text{Diagnose}(q, I, s, M, \text{TeamOutput}) $
\STATE \textbf{Return} \text{Diagnosis Result}
\end{algorithmic}
\end{algorithm}

\begin{table*}[t]
\centering
\small
\setlength{\tabcolsep}{3pt}
\caption{Performance comparison on  CheXbench (accuracy in \%). Tasks: SDI (Single Disease Identification), MDI (Multiple Disease Identification), FGR (Fine-Grained Reasoning), BDC (Binary Disease Classification), VQA (Visual Question Answering), VC (View Classification). Overall is the average over all tasks. Best results are in bold.  }
\label{tab:vqa1}
\resizebox{\textwidth}{!}{
\begin{tabular}{l*{8}{c}>{\columncolor{lightgray!15}}c>{\columncolor{lightgray!15}}c}
\toprule[1.5pt]
\multirow{2}{*}{Categories} & \multicolumn{10}{c}{Models} \\
\cmidrule(lr){2-11}
 & LLaVA-Med & CheXagent & Janus-pro-7B & GPT-4o & Qwen-VL-Max & MedGemma & MedRAX (GPT) & MedRAX (Qwen) & Ours (GPT) & Ours (Qwen) \\
\midrule[1pt]

SDI {\tiny MIMIC-CXR} & 26.7 & 30.3 & 27.7 & 29.7 & 31.7 & 38.5 & 28.8 & 32.8 & 39.3 & \textbf{42.9} \\
SDI {\tiny CheXpert}  & 26.0 & 29.6 & 27.8 & 49.7 & 51.4 & 46.1 & 34.9 & 45.6 & 52.1 & \textbf{55.6} \\

MDI {\tiny MIMIC-CXR} & 28.7 & 55.3 & 23.0 & 53.6 & 53.3 & 46.6 & 51.6 & 44.0 & \textbf{64.6} & 61.3 \\
MDI {\tiny CheXpert}  & 15.0 & 52.1 & 39.6 & 62.1 & 68.5 & 60.3 & 49.3 & 51.4 & \textbf{73.2} & 68.9 \\

VQA  {\tiny Rad-Restruct} & 34.9 & 57.3 & 51.1 & 60.8 & 58.2 & 58.2 & \textbf{68.7} & 57.4 & 62.6 & 59.1 \\
VQA  {\tiny SLAKE}    & 55.5 & 78.1 & 59.4 & 85.1 & 78.8 & 77.2 & 82.9 & 79.6 & 86.2 & \textbf{87.0} \\

FGR  {\tiny OpenI}    & 45.6 & 59.0 & 51.1 & 41.8 & 51.3 & 54.7 & 52.6 & 48.1 & 53.4 & \textbf{59.4} \\

BDC {\tiny CheXpert}  & 47.6 & \textbf{76.0} & 53.7 & 64.8 & 61.8 & 53.2 & 55.8 & 58.8 & 64.8 & 60.5 \\
BDC {\tiny SIIM}      & 49.0 & 64.0 & 43.0 & 57.0 & 51.0 & 68.0 & 67.0 & 65.0 & \textbf{77.0} & 72.0 \\
BDC {\tiny RSNA}      & 44.0 & 81.0 & 55.0 & 76.0 & 71.0 & 72.0 & 78.0 & 76.0 & \textbf{83.0} & \textbf{83.0} \\
VC  {\tiny MIMIC-CXR} & 23.8 & \textbf{97.5} & 34.0 & 79.0 & 86.6 & 77.2 & 58.7 & 65.3 & 74.0 & 87.7 \\
\midrule[1pt]
Overall               & 36.0 & 61.8 & 42.3 & 59.9 & 60.3 & 55.5 & 57.1 & 56.7 & 66.3 & \textbf{67.0} \\
\bottomrule[1.5pt]
\label{tab_vq1}
\end{tabular}
}
\end{table*}

\begin{table*}[h]
\centering
\small
\setlength{\tabcolsep}{3pt}
\caption{Performance comparison of multimodal models on Medical-CXR-VQA across clinical question types (accuracy in \%). All questions are binary classification tasks. Overall is the average over all tasks. Best results are in bold.}
\label{tab:vqa2}
\resizebox{\textwidth}{!}{
\begin{tabular}{l*{8}{c}>{\columncolor{lightgray!15}}c>{\columncolor{lightgray!15}}c}
\toprule[1.5pt]
\multirow{2}{*}{Categories} & \multicolumn{10}{c}{Models} \\
\cmidrule(l){2-11}
 & LLaVA-Med & CheXagent & Janus-pro-7B & GPT-4o & Qwen-VL-Max & MedGemma & MedRAX (GPT) & MedRAX (Qwen) & Ours (GPT) & Ours (Qwen) \\ 
\midrule[1pt]
Presence     & 50.7 & 63.3 & 59.1 & 67.2 & 63.8 & 67.8 & 67.5 & 66.4 & \textbf{70.2} & 68.0 \\
Abnormality  & 37.1 & 75.0 & 37.9 & 70.4 & 67.9 & 65.6 & 71.0 & 66.6 & 73.4 & \textbf{75.7} \\
View         & 44.4 & 48.0 & 44.0 & 70.1 & 80.6 & 81.3 & 66.0 & 78.7 & 71.4 & \textbf{83.3} \\
\midrule[1pt]
Overall      & 44.1 & 62.1 & 51.4 & 69.2 & 70.8 & 71.5 & 68.2 & 70.5 & 71.7 & \textbf{75.6} \\
\bottomrule[1.5pt]
\end{tabular}
\label{tab_vq2}
}
\end{table*}

\begin{table}[h]
\centering
\small
\setlength{\tabcolsep}{3pt}
\caption{\textcolor{black}{Performance comparison on MIMIC-CXR for report generation. $\ast$ and $\blacktriangle$ denote that CXRAgent with GPT-4o and Qwen-VL-Max, respectively, achieves statistically significant improvements over the corresponding methods ($p < 0.05$, paired Wilcoxon signed-rank test). Best results are in bold.}}
\begin{tabular}{lcccc}
\toprule[1.5pt]
Method & {LLMScore} & RadGraph & GREEN & RaTEScore \\
\midrule[1pt]
LLaVA-Med         & 1.442$^{\ast\blacktriangle}$ & 0.068$^{\ast\blacktriangle}$ & 0.014$^{\ast\blacktriangle}$ & 0.418$^{\ast\blacktriangle}$ \\
CheXagent         & 2.785$^{\ast\blacktriangle}$ & 0.246$^{\ast\blacktriangle}$ & 0.331$^{\ast}$ & 0.489$^{\ast\blacktriangle}$ \\
LLaVA-Rad         & 3.172$^{\ast\blacktriangle}$ & 0.235$^{\ast\blacktriangle}$ & 0.330$^{\ast}$ & 0.538$^{\ast\blacktriangle}$ \\
MedGemma          & 2.588$^{\ast\blacktriangle}$ & 0.135$^{\ast\blacktriangle}$ & 0.271$^{\ast\blacktriangle}$ & 0.512$^{\ast\blacktriangle}$ \\
GPT-4o            & 2.364$^{\ast\blacktriangle}$ & 0.112$^{\ast\blacktriangle}$ & 0.190$^{\ast\blacktriangle}$ & 0.480$^{\ast\blacktriangle}$ \\
Qwen-VL-Max       & 2.110$^{\ast\blacktriangle}$ & 0.119$^{\ast\blacktriangle}$ & 0.219$^{\ast\blacktriangle}$ & 0.473$^{\ast\blacktriangle}$ \\
MAIRA-2           & 2.878$^{\ast\blacktriangle}$ & 0.192$^{\ast\blacktriangle}$ & 0.304$^{\ast\blacktriangle}$ & 0.532$^{\ast\blacktriangle}$ \\
CXRMate           & 3.051$^{\ast\blacktriangle}$ & 0.213$^{\ast\blacktriangle}$ & 0.334$^{\ast}$ & 0.535$^{\ast\blacktriangle}$ \\
CXRMate-ED        & 3.077$^{\ast\blacktriangle}$ & \textbf{0.266}$^{\ast\blacktriangle}$ & 0.336$^{\ast}$ & 0.540$^{\ast\blacktriangle}$ \\
CvT2DistilGPT2    & 2.733$^{\ast\blacktriangle}$ & 0.219$^{\ast\blacktriangle}$ & 0.318$^{\ast\blacktriangle}$ & 0.519$^{\ast\blacktriangle}$ \\
MedVersa          & 2.066$^{\ast\blacktriangle}$ & 0.056$^{\ast\blacktriangle}$ & 0.094$^{\ast\blacktriangle}$ & 0.449$^{\ast\blacktriangle}$ \\
IFCC              & 2.723$^{\ast\blacktriangle}$ & 0.265$^{\ast\blacktriangle}$ & 0.356$^{\ast\blacktriangle}$ & 0.568$^{\ast\blacktriangle}$ \\
Libra             & 2.942$^{\ast\blacktriangle}$ & 0.144$^{\ast\blacktriangle}$ & 0.299$^{\ast\blacktriangle}$ & 0.548$^{\ast\blacktriangle}$ \\
MedRAX (GPT)         & 2.388$^{\ast\blacktriangle}$ & 0.122$^{\ast\blacktriangle}$ & 0.201$^{\ast\blacktriangle}$ & 0.486$^{\ast\blacktriangle}$ \\
MedRAX (Qwen)         & 2.267$^{\ast\blacktriangle}$ & 0.126$^{\ast\blacktriangle}$ & 0.220$^{\ast\blacktriangle}$ & 0.478$^{\ast\blacktriangle}$ \\
\midrule[1pt]
Ours (GPT)           & 3.267 & 0.251 & \textbf{0.359} & \textbf{0.569} \\
Ours (Qwen)           & \textbf{3.337} & 0.243 & 0.334 & 0.557 \\
\bottomrule[1.5pt]
\label{com_table_report}
\end{tabular}
\end{table}

\section{Experiments}
\subsection{Experiment Setup}
\textbf{Datasets and Tasks.}  
We evaluate CXRAgent across multiple CXR interpretation tasks on the following three datasets: (1) CheXbench \cite{chen2024chexagent} serves as a comprehensive benchmark crafted to assess three aspects of CXR analysis including image perception, image-text reasoning, and text generation. We focus on image perception (i.e., view classification and disease identification), image-text reasoning (fine-grained reasoning, visual question answering (VQA)). (2) Medical-CXR-VQA \cite{HU2024103279} is a benchmark for CXR visual question answering. Following its official tasks, we randomly sample 594 Presence, 256 Abnormality, and 150 View questions to  assess the model. (3) MIMIC-CXR \cite{MIMIC} provides 377,110 CXRs paired with 227,835 reports. \textcolor{black}{We employ the entire testing split for the evaluation of CXR report generation. The data are available in our project for reproducibility. }

\noindent\textbf{Implementation Details.} We use GPT-4o or Qwen-VL-Max as the core director. \textcolor{black}{For brevity, we refer to them as GPT and Qwen in the following tables.} CXRAgent is deployed on two NVIDIA RTX A6000 GPUs. \textcolor{black}{We use the low-temperature decoding strategy during inference to reduce random fluctuations across runs.}  To enhance diagnostic capabilities of our CXRAgent, we have integrated a set of advanced tools:
\begin{itemize}
    \item \textbf{MedGemma-4B \cite{medgemma-hf}:} An open-source medical multimodal model finetuned on Gemma3, featuring a  SigLIP vision encoder for clinical imaging analysis and visual-language reasoning.
    \item \textbf{LLaVA-Rad \cite{zambrano2025clinically}:} A multimodal model, optimized for automated radiology report generation from CXR images.  
    \item \textbf{CheXagent \cite{chen2024chexagent}:} A foundation model for interpretation of chest X-rays (e.g., disease classification and localization). 
    \item \textbf{LLaVA-Med \cite{li2023llava}:} A multimodal assistant enabling interactive image-based medical diagnosis and conversational reasoning over radiological findings.
    \item \textbf{MAIRA-2 \cite{bannur2024maira}:} A radiology-specific multimodal model for grounded report generation, ensuring that textual descriptions are spatially anchored to the image.
    \item \textbf{MedVLM-R1 \cite{pan2025medvlm}:} A vision-language model for radiological tasks that generates natural language reasoning alongside the final answer.
\end{itemize}
\textcolor{black}{By orchestrating these heterogeneous tools, the Director effectively leverages their complementary strengths across diverse tasks. Importantly, this modular, plug-and-play architecture allows seamless integration of newer and more advanced models, further enhancing the agent's diagnostic performance.}

\noindent\textbf{Baselines.} We compare our framework with three representative types of baseline methods: \textcolor{black}{1) general-purpose multimodal LLMs (Qwen-VL-Max, GPT-4o, Janus-pro-7B) with cross-domain medical adaptation capabilities; 2) radiology-specialized models (LLaVA-Med, CheXagent, LLaVA-Rad, MedGemma, MAIRA-2, CXRMate, CXRMate-ED, CvT2DistilGPT2, MedVersa, IFCC, Libra); and 3) MedRAX, an agent that integrates various diagnostic tools for CXR analysis without   fine-tuning.}  

\noindent\textbf{Evaluation Metrics. } 

For report generation, we use four metrics: 1) RaTEScore \cite{zhao2024ratescore} with entity-level evaluation of  reports by assessing critical elements through its specialized named entity recognition (NER) framework; 2) LLMScore, implemented with GPT-4o, which rates reports on a 1–5 scale following the ITU five-point standard (1 = Bad, 5 = Excellent); \textcolor{black}{3) GREEN \cite{2024-green}, an interpretable and open-source LLM-based evaluation pipeline; 4) RadGraph-F1 \cite{delbrouck-etal-2022-improving}, judging the correctness of entities and relations.  We calculate RaTEScore, GREEN and RadGraph-F1 using the toolkit \cite{radeval}.} For other tasks, we employ accuracy for assessment. Results are reported as the average over all samples.

\begin{table*}[!ht]
\centering
\small
\setlength{\tabcolsep}{2pt}
\renewcommand{\arraystretch}{1.2}
\caption{\textcolor{black}{Performance comparison of CXR report generation on 200 simple and 200 complex cases. Best results are in bold.}}
\label{tab:report_simple_complex_3decimal}
\resizebox{\textwidth}{!}{
\begin{tabular}{lccccccccccccccccc}
\toprule[1.5pt]
\textbf{Metrics} & \textbf{LLaVA-Med} & \textbf{CheXagent} & \textbf{LLaVA-Rad} & \textbf{MedGemma} & \textbf{GPT-4o} & \textbf{Qwen-VL-Max} & \textbf{MAIRA-2} & \textbf{CXRMate} & \textbf{CXRMate-ED} & \textbf{CvT2DistilGPT2} & \textbf{MedVersa} & \textbf{IFCC} & \textbf{Libra} & \textbf{MedRAX (GPT)} & \textbf{MedRAX (Qwen)} & \textbf{Ours (GPT)} & \textbf{Ours (Qwen)} \\
\midrule[1pt]
LLMScore & 1.435 & 3.454 & 3.512 & 2.587 & 2.546 & 2.450 & 3.246 & 3.354 & 3.543 & 3.431 & 2.130 & 3.183 & 3.082 & 2.675 & 2.743 & \textbf{3.564} & 3.545 \\
RadGraph & 0.065 & 0.323 & 0.293 & 0.187 & 0.144 & 0.155 & 0.270 & 0.288 & 0.324 & 0.311 & 0.029 & \textbf{0.346} & 0.260 & 0.156 & 0.163 & 0.307 & 0.285 \\
GREEN & 0.005 & 0.504 & 0.482 & \textbf{0.559} & 0.484 & 0.538 & 0.533 & \textbf{0.570} & 0.500 & 0.558 & 0.025 & 0.554 & 0.437 & 0.484 & 0.526 & 0.532 & 0.511 \\
RaTEScore & 0.402 & 0.604 & 0.611 & 0.526 & 0.506 & 0.505 & 0.601 & 0.587 & 0.618 & 0.603 & 0.444 & \textbf{0.628} & 0.590 & 0.517 & 0.513 & 0.627 & 0.606 \\
\midrule[1pt]
LLMScore & 1.256 & 2.551 & 2.856 & 2.578 & 2.122 & 1.980 & 2.487 & 2.795 & 2.937 & 2.074 & 2.148 & 3.022 & 2.798 & 2.256 & 2.245 & \textbf{3.208} & 3.187 \\
RadGraph & 0.067 & 0.184 & 0.180 & 0.099 & 0.089 & 0.090 & 0.149 & 0.159 & \textbf{0.218} & 0.147 & 0.057 & 0.212 & 0.186 & 0.098 & 0.094 & 0.189 & 0.185 \\
GREEN & 0.020 & 0.234 & 0.259 & 0.177 & 0.092 & 0.095 & 0.219 & 0.224 & 0.244 & 0.186 & 0.121 & 0.258 & 0.236 & 0.101 & 0.100 & \textbf{0.267} & 0.245 \\
RaTEScore & 0.402 & 0.509 & 0.534 & 0.512 & 0.472 & 0.470 & 0.500 & 0.492 & 0.549 & 0.457 & 0.456 & 0.530 & 0.525 & 0.476 & 0.472 & \textbf{0.554} & 0.539 \\
\bottomrule[1.5pt]
\label{Complex}
\end{tabular}
}
\end{table*}

\subsection{Performance Comparison}

\textbf{CheXbench.} As demonstrated in Table~\ref{tab:vqa1}, our Qwen-based CXRAgent achieves state-of-the-art performance on the comprehensive CheXbench benchmark, attaining an overall accuracy of 67.0\%. The framework demonstrates remarkable improvements across multiple challenging tasks, particularly excelling in complex diagnostic scenarios. For multi-disease identification, CXRAgent achieves 73.2\% on CheXpert and 64.6\% on MIMIC-CXR, representing significant advancements over previous methods. The model's robustness is especially evident in fine-grained reasoning tasks, where it reaches 59.4\% accuracy on OpenI, outperforming all comparable approaches. This consistent performance across diverse evaluation categories underscores the effectiveness of our multi-stage adaptive design, particularly the EDV module's capability to resolve tool inconsistencies through evidence-based validation. 

\textbf{Medical-CXR-VQA.} As shown in Table~\ref{tab:vqa2}, CXRAgent delivers exceptional performance on the Medical-CXR-VQA benchmark, achieving a leading overall accuracy of 75.6\%. The framework exhibits particularly strong capabilities in abnormality detection (75.7\%) and view classification (83.3\%), demonstrating its proficiency in handling clinically nuanced questioning. Notably, our approach maintains superior performance across all question types, with presence detection reaching 70.2\% accuracy. The adaptive team recruitment mechanism proves crucial in these visual question answering scenarios, enabling dynamic collaboration patterns that effectively address the varied complexity of clinical inquiries. The superiority across both benchmarks demonstrates CXRAgent’s strong adaptability.  \textcolor{black}{On both the CheXbench and  Medical-CXR-VQA datasets, we conduct significance tests with McNemar's test for the accuracy over all questions. The results show that our method, using either GPT-4o or Qwen-VL-Max,  outperforms all other methods with the p-value (\textbf{$p < 0.05$}).}

 \textcolor{black}{\textbf{MIMIC-CXR.} As demonstrated in Table~\ref{com_table_report},  CXRAgent with GPT-4o   performs  well across all four evaluation metrics, especially in RaTEScore and GREEN, where it achieves scores of 0.569 and 0.359, outperforming other models and demonstrating the best performance. Compared with other methods, CXRAgent with GPT-4o also shows strong competitiveness in LLMScore and RadGraph, with scores of 3.267 and 0.251, indicating its excellent performance. While CXRAgent with Qwen-VL-Max  slightly lags behind CXRAgent with GPT-4o  in RaTEScore and GREEN, it maintains a high LLMScore reaching 3.337, showing the robustness of our method. We also conducted paired Wilcoxon signed-rank test. Results in Table~\ref{com_table_report} indicate that CXRAgent with GPT-4o or Qwen-VL-Max as the director outperforms most of the other methods across metrics, with statistical significance of $p < 0.05$. This demonstrates that the observed performance is not random.} 
 
 \textcolor{black}{To evaluate our model's performance in complex clinical scenarios, we further analyzed the distribution of reports in the testing set of MIMIC-CXR. We utilized GPT-4o to perform an ambiguity and complexity assessment across all 3,858 reports in the test set and all samples were ranked. We specifically define complex cases as those involving intricate anatomical structures, post-surgical hardware, or ambiguous diagnostic conclusions that pose significant challenges. We have curated 200 of the most complex samples and 200 of the simplest samples for report generation to facilitate performance comparison. Results are shown in Table \ref{Complex}, revealing several key insights.  First, as report complexity increases, all models exhibit a performance decline across various metrics, which is consistent with the increased difficulty of interpreting multi-finding clinical cases. However, our CXRAgent demonstrates superior resilience compared with other baselines. In terms of the LLMScore, which reflects clinical logic and diagnostic accuracy, our CXRAgent with GPT-4o maintains a leading score of 3.208 in complex scenarios, whereas many other models experience much sharper drops when faced with high-density medical jargon and diagnostic uncertainty. Furthermore, our model achieves the highest RaTEScore of 0.554 in the  200 complex subset,  outperforming SOTA models like  CXRMate-ED (0.549),  LLaVA-Rad (0.534) and IFCC (0.530). This indicates that our model is more effective at capturing critical clinical facts in challenging scenarios involving multiple diseases or subtle differential diagnoses. This performance advantage is primarily attributed to our Director-Orchestrated Multi-Stage Reasoning framework and the EDV module.}

\subsection{Ablation Study}

\textbf{CheXbench}. Shown in Tables~\ref{tab:ablation1},~\ref{tab:ablation2}, the ablation results reveal three key findings:  First, although tool integration generally improves performance, incorporating unverified tool outputs can occasionally degrade the performance. For example, in the view classification task on CheXbench (Qwen), accuracy drops from 86.6\% without tools to 83.6\% with tools.  Second, the EDV module effectively mitigates such errors by resolving conflicts through visual evidence validation, playing a critical role in ensuring reliability when tool outputs are inconsistent. Third, the full model that combines tools, EDV, and team collaboration delivers optimal performance across tasks, highlighting the synergistic contribution of all components. With GPT-4o as the director, it yields a relative accuracy improvement of 10.7\% over the baseline and shows particular advantages in complex reasoning tasks, increasing multi-disease identification (CheXpert) to 73.2\% and fine-grained interpretation (OpenI) to 53.4\%. Overall, these results demonstrate that CXRAgent delivers robust and consistent gains across different large language model backbones, confirming that the combination of specialized tools, evidence-driven validation, and collaborative reasoning leads to clinically reliable and generalizable diagnostic performance.

\textbf{Medical-CXR-VQA}. The ablation results presented in Table~\ref{tab:ablation3} demonstrate consistent performance improvements through the progressive integration of our framework's components on the Medical-CXR-VQA benchmark. Three key patterns emerge from the analysis. First, the incorporation of specialized CXR-analysis tools establishes a solid foundation for accurate diagnosis, with the Qwen-directed system showing particular strength in abnormality detection (increasing from 67.9\% to 73.4\%). Second, the EDV module contributes significantly to diagnostic reliability, particularly for presence detection tasks where the GPT-4o-based system improves from 68.0\% to 70.0\%. This enhancement confirms EDV's role in validating tool outputs against visual evidence, reducing diagnostic uncertainty. Third, the complete framework achieves optimal performance across all question types, with the Qwen-based configuration reaching 75.6\% overall accuracy, representing a 4.8\% absolute improvement over the baseline. The framework demonstrates robust performance across different director models, with both Qwen and GPT-4o configurations showing the highest scores in their respective categories. These results validate that the synergistic combination of specialized tools, evidence-based validation, and collaborative reasoning enables more reliable and comprehensive CXR interpretation.

\textbf{MIMIC-CXR}. Table~\ref{tab:ablation4} presents the ablation results for report generation. While integrating tools alone enhances diagnostic accuracy, the full framework integrating both EDV validation and team collaboration yields the best performance. It reveals three critical findings about our framework’s report generation capability. First, integrating specialized CXR-analysis tools substantially enhances diagnostic performance, confirming their value in capturing domain-specific patterns. Second, the EDV module not only improves diagnostic accuracy but also impacts raw entity extraction, enhancing clinical report quality by enforcing evidence-backed coherence. Third, the complete system achieves optimal performance through collaborative reasoning, where virtual specialist teams demonstrate unique value in synthesizing comprehensive findings beyond the capability of individual tools. Overall, these results validate our core design: combining specialized tools with evidence-based validation and expert team coordination produces the most clinically reliable reports, highlighting our framework’s strength in generating clinically grounded  diagnostic reports rather than merely listing observations.

\begin{table}[t]
\centering
\small
\setlength{\tabcolsep}{4pt}
\caption{Ablation study on CheXbench (accuracy in \%, with Qwen as director).  Best results are in bold. For abbreviation definitions, please refer to Table 1. }
\label{tab:ablation1}
\begin{tabular}{l*{4}{c}}
\toprule[1.5pt]
\multirow{2}{*}{Evaluation Task} & \multicolumn{4}{c}{Component Configuration} \\
\cmidrule(lr){2-5}
 & None & Tools & Tools+EDV & \cellcolor{lightgray!15}Full Model \\
\midrule[1pt]
SDI {\tiny MIMIC-CXR}& 31.7 & 35.3 & 40.5 & \cellcolor{lightgray!15}\textbf{42.9} \\
SDI {\tiny CheXpert} & 51.4 & 52.1 & 54.4 & \cellcolor{lightgray!15}\textbf{55.6} \\
MDI {\tiny MIMIC-CXR}& 53.3 & 55.3 & 58.6 & \cellcolor{lightgray!15}\textbf{61.3} \\
MDI {\tiny CheXpert} & 68.5 & 66.4 & 67.5 & \cellcolor{lightgray!15}\textbf{68.9} \\
VQA  {\tiny Rad-Restruct} & 58.2 & 57.3 & \textbf{59.1} &\cellcolor{lightgray!15} \textbf{59.1} \\
VQA  {\tiny SLAKE} & 78.8 & 84.5 & 86.1 &\cellcolor{lightgray!15} \textbf{87.0} \\
FGR {\tiny OpenI} & 51.3 & 52.3 & 53.4 & \cellcolor{lightgray!15}\textbf{59.4} \\
BDC {\tiny CheXpert} & 61.8 & \textbf{62.2} & 61.8 & \cellcolor{lightgray!15}60.5 \\
BDC {\tiny SIIM} & 51.0 & 68.0 & 71.0 & \cellcolor{lightgray!15}\textbf{72.0} \\
BDC {\tiny RSNA} & 71.0 & 76.0 & 80.0 & \cellcolor{lightgray!15}\textbf{83.0} \\
VC {\tiny MIMIC-CXR} & 86.6 & 83.6 & 86.6 & \cellcolor{lightgray!15}\textbf{87.7} \\
\midrule[1pt]
Overall & 60.3 & 63.0 & 65.4 & \cellcolor{lightgray!15}\textbf{67.0} \\
\bottomrule[1.5pt]
\end{tabular}
\end{table}

\begin{table}[h]
\centering
\small
\setlength{\tabcolsep}{4pt}
\caption{Ablation study on CheXbench (accuracy in \%, with GPT-4o  as director).  Best results are in bold.} 
\label{tab:ablation2}
\begin{tabular}{l*{4}{c}}
\toprule[1.5pt]
\multirow{2}{*}{Evaluation Task} & \multicolumn{4}{c}{Component Configuration} \\
\cmidrule(lr){2-5}
 & None & Tools & Tools+EDV & \cellcolor{lightgray!15}Full Model \\
\midrule[1pt]
SDI {\tiny MIMIC-CXR}& 29.7 & 33.3 & 36.9 & \cellcolor{lightgray!15}\textbf{39.3} \\
SDI {\tiny CheXpert} & 49.7 & 50.3 & 51.5 & \cellcolor{lightgray!15}\textbf{52.1} \\
MDI {\tiny MIMIC-CXR}& 53.6 & 56.7 & 61.7 & \cellcolor{lightgray!15}\textbf{64.6} \\
MDI {\tiny CheXpert} & 62.1 & 59.6 & 65.4 & \cellcolor{lightgray!15}\textbf{73.2} \\
VQA  {\tiny Rad-Restruct} & 60.8 & 55.7 & \textbf{66.1} &\cellcolor{lightgray!15}62.6 \\
VQA  {\tiny SLAKE} & 85.1 & 82.9 & 84.6 &\cellcolor{lightgray!15}\textbf{86.2} \\
FGR {\tiny OpenI} & 41.8 & 45.3 & 47.9 & \cellcolor{lightgray!15}\textbf{53.4} \\
BDC {\tiny CheXpert} & \textbf{64.8} & 64.0 & 65.2 & \cellcolor{lightgray!15}\textbf{64.8} \\
BDC {\tiny SIIM} & 57.0 &66.0 & 74.0 & \cellcolor{lightgray!15}\textbf{77.0} \\
BDC {\tiny RSNA} & 76.0 & 79.0 & 81.0 & \cellcolor{lightgray!15}\textbf{83.0} \\
VC {\tiny MIMIC-CXR} & \textbf{79.0} & 72.7 & 73.3 & \cellcolor{lightgray!15}74.0 \\
\midrule[1pt]
Overall &59.9& 60.5 & 64.3 & \cellcolor{lightgray!15}\textbf{66.3} \\
\bottomrule[1.5pt]
\end{tabular}
\end{table}

\begin{table}[t]
\centering
\small
\setlength{\tabcolsep}{4pt}
\caption{Ablation study on Medical-CXR-VQA (accuracy in \%) under directors (Qwen vs. GPT-4o). Best results are in bold.}
\label{tab:ablation3}
\begin{tabular}{l*{4}{c}}
\toprule[1.5pt]
\multirow{2}{*}{Evaluation Task} & \multicolumn{4}{c}{Component Configuration} \\
\cmidrule(lr){2-5}
 & None & Tools & Tools+EDV & \cellcolor{lightgray!15}Full Model \\
\midrule[1pt]
\rowcolor{gray!10}
\multicolumn{5}{l}{\textbf{(a) Using Qwen-VL-Max as Director}} \\
Presence & 63.8 & 66.3 & 66.6 & \cellcolor{lightgray!15}\textbf{68.0} \\
Abnormality & 67.9 & 73.4 & 75.3 & \cellcolor{lightgray!15}\textbf{75.7} \\
View & 80.6 & 80.0 & 82.0 & \cellcolor{lightgray!15}\textbf{83.3} \\
\midrule[0.5pt]
\textbf{Overall (Qwen)} & 70.8 & 73.2 & 74.6 & \cellcolor{lightgray!15}\textbf{75.6} \\
\midrule[0.75pt]
\rowcolor{gray!10}
\multicolumn{5}{l}{\textbf{(b) Using GPT-4o as Director}} \\
Presence & 67.2 & 68.0 & 70.0 & \cellcolor{lightgray!15}\textbf{70.2} \\
Abnormality & 70.4 & 71.8 & 72.7 & \cellcolor{lightgray!15}\textbf{73.4} \\
View & 70.1 & 69.3 & 70.7 & \cellcolor{lightgray!15}\textbf{71.4} \\
\midrule[0.5pt]
\textbf{Overall (GPT)} & 69.2 & 69.7 & 71.1 & \cellcolor{lightgray!15}\textbf{71.7} \\
\bottomrule[1.5pt]
\end{tabular}
\end{table}

\begin{table}[h]
\centering
\small
\caption{\textcolor{black}{Ablation study on report generation (best results in bold).  "None" is the baseline model; "Tools" indicates the inclusion of tools; "Tools+EDV" integrates tools and EDV; "Full" is the complete model with all proposed components.}}
\label{tab:ablation_all}
\begin{tabular}{llccc>{\columncolor{lightgray!20}}c}
\toprule[1.5pt]
Director & Metric & None & Tools & Tools+EDV & \textbf{Full} \\
\midrule[1pt]
\multirow{4}{*}{GPT} 
& LLMScore  & 2.364 & 2.837 & 3.203 & \textbf{3.267} \\
& GREEN     & 0.190 & 0.322 & 0.345 & \textbf{0.359} \\
& RaTEScore & 0.480 & 0.550 & 0.560 & \textbf{0.569} \\
& RadGraph  & 0.112 & 0.221 & 0.231 & \textbf{0.251} \\
\midrule
\multirow{4}{*}{Qwen}
& LLMScore  & 2.110 & 3.066 & 3.310 & \textbf{3.337} \\
& GREEN     & 0.219 & 0.326 & 0.331 & \textbf{0.334} \\
& RaTEScore & 0.473 & 0.521 & 0.553 & \textbf{0.557} \\
& RadGraph  & 0.119 & 0.207 & 0.215 & \textbf{0.243} \\
\bottomrule[1.5pt]
\label{tab:ablation4}
\end{tabular}
\end{table}

\subsection{Analysis of EDV and Memory}

\textcolor{black}{To quantify the effectiveness of EDV with misleading and conflicting conditions, we designed a controlled experiment using the Single Disease Identification (SDI) task from CheXbench. It contains data from two datasets (MIMIC-CXR \cite{MIMIC} and CheXpert \cite{irvin2019chexpert}). We simulated a high-conflict scenario by treating all candidate options as independent tool outputs. In this setup, only one output is correct while the remaining three serve as misleading inputs, creating a challenging 3:1 conflict ratio that mimics real-world tool noise. We evaluated two configurations: (1) \textit{w/o EDV}, where all candidate outputs are directly aggregated; and (2) \textit{w/ EDV}, where each candidate is validated through evidence grounding and confidence estimation before final decision-making. As shown in Table~\ref{tab:edv_trust}, EDV consistently improves QA accuracy across datasets and LLM backbones under this high-conflicting setting. For example, on MIMIC-CXR, the Qwen-based model improves from 26.1\% to 31.8\% (+5.7\%), while on CheXpert, the GPT-based model improves from 42.0\% to 46.2\% (+4.2\%). These results demonstrate that EDV effectively suppresses unreliable or weakly supported conclusions and prioritizes evidence-backed predictions. }

\textcolor{black}{To further evaluate EDV on diagnostically ambiguous and complex cases, we curated a subset of the 200 most complex cases to evaluate the robustness of EDV in scenarios where underlying tools are most prone to failure for report generation. The sample selection is  described in the section ``Performance Comparison" of MIMIC-CXR. As shown in Table \ref{tab_200_comp}, the integration of EDV consistently yields improvements in clinical consistency metrics. Specifically, in these 200 complex cases, the RadGraph score increases from 0.177 to 0.188 in the GPT-4o configuration, and RaTEScore improves from 0.512 to 0.536 on Qwen-VL-Max setup. This indicates that EDV effectively identifies erroneous findings in the complex challenging scenarios.}

\begin{table}[h]
\centering
\small
\caption{\textcolor{black}{Evaluation of EDV under misleading tool outputs on Single Disease Identification of CheXbench  (accuracy in \%).}}
\begin{tabular}{l l c c c}
\toprule
Dataset & LLM & w/o EDV & w/ EDV & $\Delta$ \\
\midrule
\multirow{2}{*}{MIMIC-CXR}
& GPT-4o   & 25.6 & \textbf{27.7} & +2.1 \\
&  Qwen-VL-Max  & 26.1 & \textbf{31.8} & +5.7 \\
\midrule
\multirow{2}{*}{CheXpert}
& GPT-4o   & 42.0 & \textbf{46.2} & +4.2 \\
&  Qwen-VL-Max & 43.7 & \textbf{45.6} & +1.9 \\
\bottomrule
\end{tabular}
\label{tab:edv_trust}
\end{table}

\begin{table}[!ht]
\centering
\small
\setlength{\tabcolsep}{3pt}  
\caption{\textcolor{black}{Ablation study on 200 complex cases for report generation.}}
\label{tab:ablation_combined_complex}
\begin{tabular}{l*{6}{c}}
\toprule[1.5pt]
\multirow{2}{*}{Metrics} 
& \multicolumn{3}{c}{CXRAgent  (GPT-4o)} 
& \multicolumn{3}{c}{CXRAgent  (Qwen-VL-Max)} \\
\cmidrule(lr){2-4} \cmidrule(lr){5-7}
& None & Tools & Tools+EDV 
& None & Tools & Tools+EDV \\
\midrule[1pt]
LLMScore  
& 2.122 & 3.128 & \textbf{3.198} 
& 1.980 & 2.945 & \textbf{3.028} \\

RaTEScore 
& 0.472 & 0.531 & \textbf{0.544} 
& 0.470 & 0.512 & \textbf{0.536} \\

GREEN 
& 0.092 & 0.231 & \textbf{0.266} 
& 0.095 & 0.212 & \textbf{0.239} \\

RadGraph  
& 0.089 & 0.177 & \textbf{0.188} 
& 0.090 & 0.177 & \textbf{0.179} \\
\bottomrule[1.5pt]
\label{tab_200_comp}
\end{tabular}
\end{table}

\textcolor{black}{To quantify importance of the memory mechanism, we conducted an ablation study comparing CXRAgent with a ``Memory-less" variant. Results are shown in Table \ref{tab:memory_ablation_en}. In the memory-less setup, the Director is forced to make decisions  without accessing the conclusions of previously invoked tools or  EDV's validation results when forming a diagnostic plan or assembling teams. The performance of report generation drops sharply across both GPT-4o and  Qwen-VL-Max backbones when memory is removed. For instance,  RaTEScore decreases from 0.569 to 0.538 (GPT) and from 0.557 to 0.547 (Qwen). This confirms that the accumulation of reasoning traces in memory is essential for the Director to maintain diagnostic continuity and rectify errors in complex clinical scenarios.}
\begin{table}[h]
\centering
\small
\setlength{\tabcolsep}{6pt}
\caption{\textcolor{black}{Ablation study of Memory for report generation on MIMIC-CXR. ``Mem." indicates  memory. }}
\begin{tabular}{lcccc}
\toprule[1.5pt]
\multirow{2}{*}{Metrics} & \multicolumn{2}{c}{Ours  (GPT-4o)} & \multicolumn{2}{c}{Ours (Qwen-VL-Max)} \\
\cmidrule(lr){2-3} \cmidrule(lr){4-5}
 & w/o Mem. & w/ Mem. & w/o Mem. & w/ Mem. \\
\midrule[1pt]
LLMScore  & 3.018 & \textbf{3.267} & 2.982 & \textbf{3.337} \\
RadGraph   & 0.226 & \textbf{0.251} & 0.204 & \textbf{0.243} \\
GREEN      & 0.336 & \textbf{0.359} & 0.302 & \textbf{0.334} \\
RaTEScore  & 0.538 & \textbf{0.569} & 0.547 & \textbf{0.557} \\
\bottomrule[1.5pt]
\label{tab:memory_ablation_en}
\end{tabular}
\end{table}

\subsection{Case Study}

\begin{figure*}[h]
\centering
\includegraphics[width=1\textwidth]{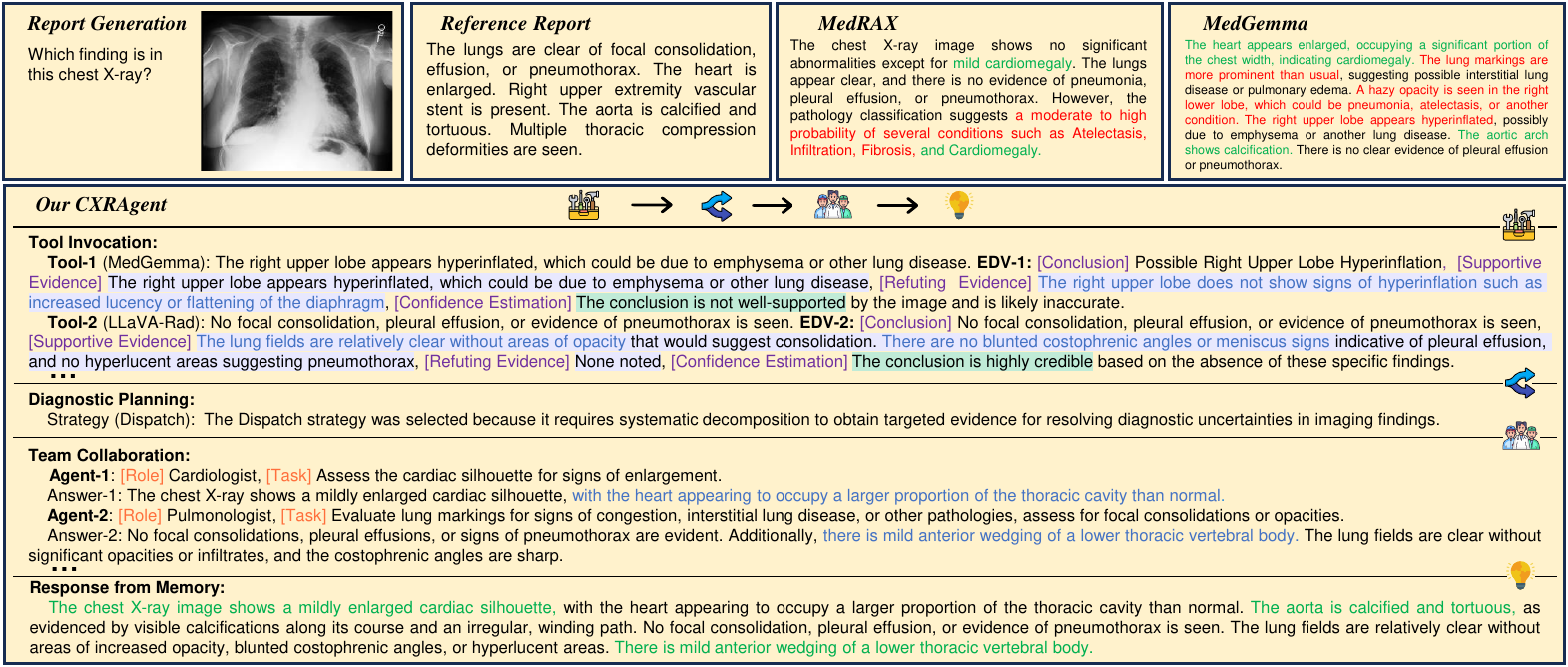}
\caption{Case study of CXRAgent versus state-of-the-art methods on report generation. \textbf{Top}: user query, reference report, and baseline predictions. \textbf{Bottom}: Prediction of CXRAgent. EDV sections are highlighted in purple with key evidence in blue and confidence highlighted in green. The roles and tasks of team members are in orange. Correct findings are in green and incorrect ones in red.}
\label{fig3}
\end{figure*}

\textbf{Case1.} As illustrated in Fig.~\ref{fig3}, CXRAgent performs multi-stage reasoning for chest X-ray interpretation, demonstrating strong capability in handling complex diagnostic cases through evidence-based validation and adaptive collaboration. While baseline models such as MedRAX and MedGemma produce conflicting or partially inaccurate findings, CXRAgent ensures diagnostic consistency through systematic reasoning.

The process begins with the invocation of two specialized tools: MedGemma and LLaVA-Rad. MedGemma provides a detailed yet over-sensitive analysis, reporting abnormalities such as “right upper lobe hyperinflation” and “hazy opacity in the right lower lobe,” whereas LLaVA-Rad offers a more conservative assessment, indicating no focal consolidation, pleural effusion, or pneumothorax.

To verify these outputs, the EDV module conducts visual grounding. For MedGemma’s hyperinflation claim, EDV detects no key radiographic signs such as increased lucency or diaphragmatic flattening and thus assigns a “not well-supported” rating. In contrast, LLaVA-Rad’s negative findings are validated with high confidence, supported by clear lung fields and sharp costophrenic angles. Given the conflicting evidence, CXRAgent initiates diagnostic planning and adopts the \textit{Dispatch} strategy, assembling a virtual multidisciplinary team of a cardiologist and pulmonologist for  evaluation. The final synthesis stage integrates validated evidence, expert analyses, and prior reasoning trajectories, effectively resolving initial tool conflicts, discarding unsupported findings, and confirming true abnormalities, including the novel vertebral wedging observation. The resulting report achieves superior clinical accuracy compared with all baselines, highlighting CXRAgent’s ability to convert diverse tool outputs into coherent, evidence-grounded, and  reliable diagnostic conclusions.

\begin{figure*}[h]
\centering
\includegraphics[width=1\textwidth]{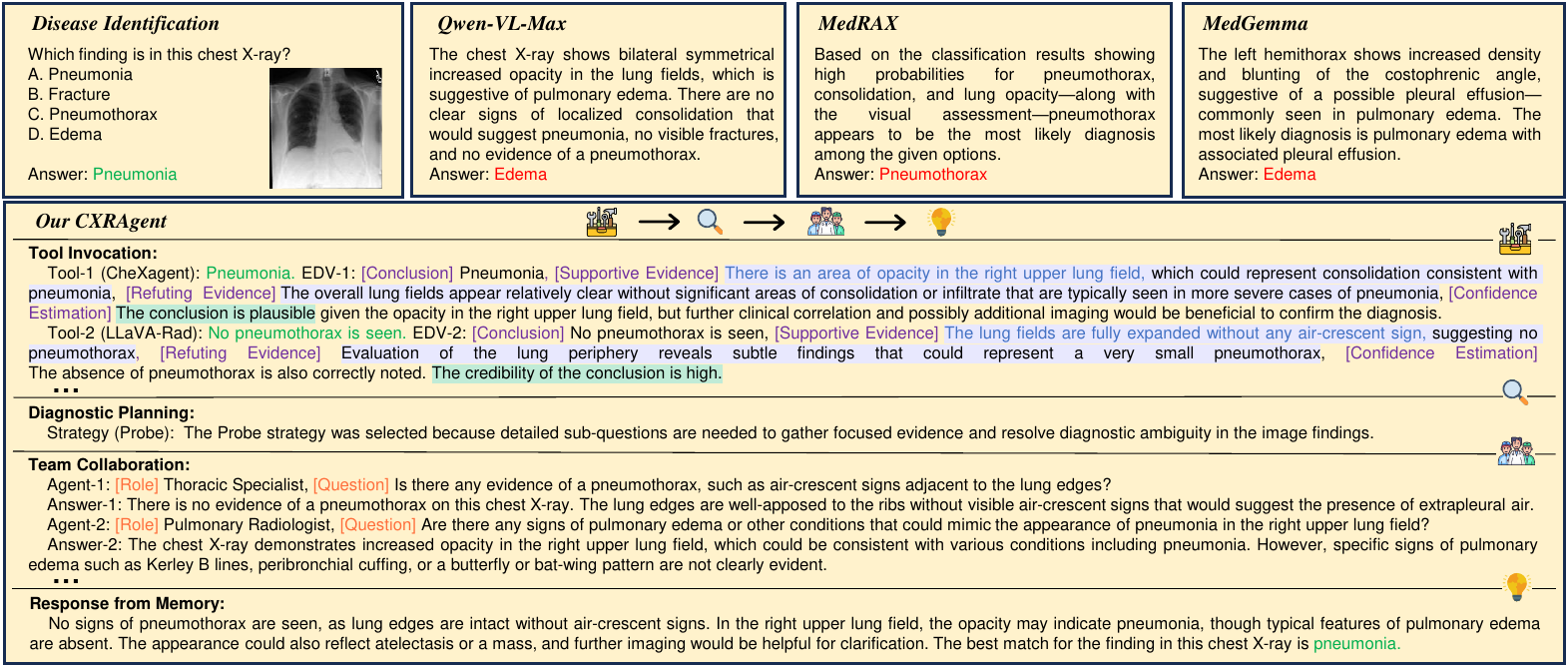}
\caption{Case study of CXRAgent versus state-of-the-art methods on visual question answer. \textbf{Top}: question, and baseline predictions. \textbf{Bottom}: Prediction of CXRAgent. EDV sections are highlighted in purple with key evidence in blue and confidence highlighted in green. The roles and tasks of team members are in orange. Correct results are in green and incorrect ones in red.}
\label{fig4}
\end{figure*}

\textbf{Case2. }As shown in Fig.~\ref{fig4}, CXRAgent demonstrates superior capabilities through its multi-stage reasoning framework when analyzing this challenging chest X-ray case. While Qwen-VL-Max, MedRAX, and MedGemma all provided incorrect diagnoses, CXRAgent correctly identified pneumonia through its evidence-based multi-stage reasoning.

In the tool invocation, CXRAgent intelligently coordinated multiple specialized tools while critically evaluating their outputs through the Evidence-driven Validator, performing comprehensive bidirectional validation. For example, it deemed CheXagent’s diagnosis ``pneumonia" was plausible based on the right upper lobe opacity, while also recognizing that the statement pertains to a localized assessment, as suggested by the remark ``the overall lung fields appear relatively clear". Such balanced evaluation differs from other models that produce erroneous outputs without evidence-backed verification.

Also, CXRAgent's diagnostic planning showed good adaptability. Recognizing the ambiguous findings, it strategically selected the \textit{Probe} collaboration mode. This allowed targeted investigation of critical differential diagnoses through specialized virtual teams. The thoracic specialist definitively ruled out pneumothorax by confirming well-apposed lung edges, while the pulmonary radiologist excluded edema by noting the absence of characteristic signs.

In addition, CXRAgent integrates the memory context for diagnosis, enabling comprehensive synthesis of findings. By maintaining and referencing prior tool validations and expert analyses, CXRAgent could weigh all evidence to determine the most probable diagnosis while transparently acknowledging diagnostic limitations.

\begin{figure*}[h]
\centering
\includegraphics[width=1\textwidth]{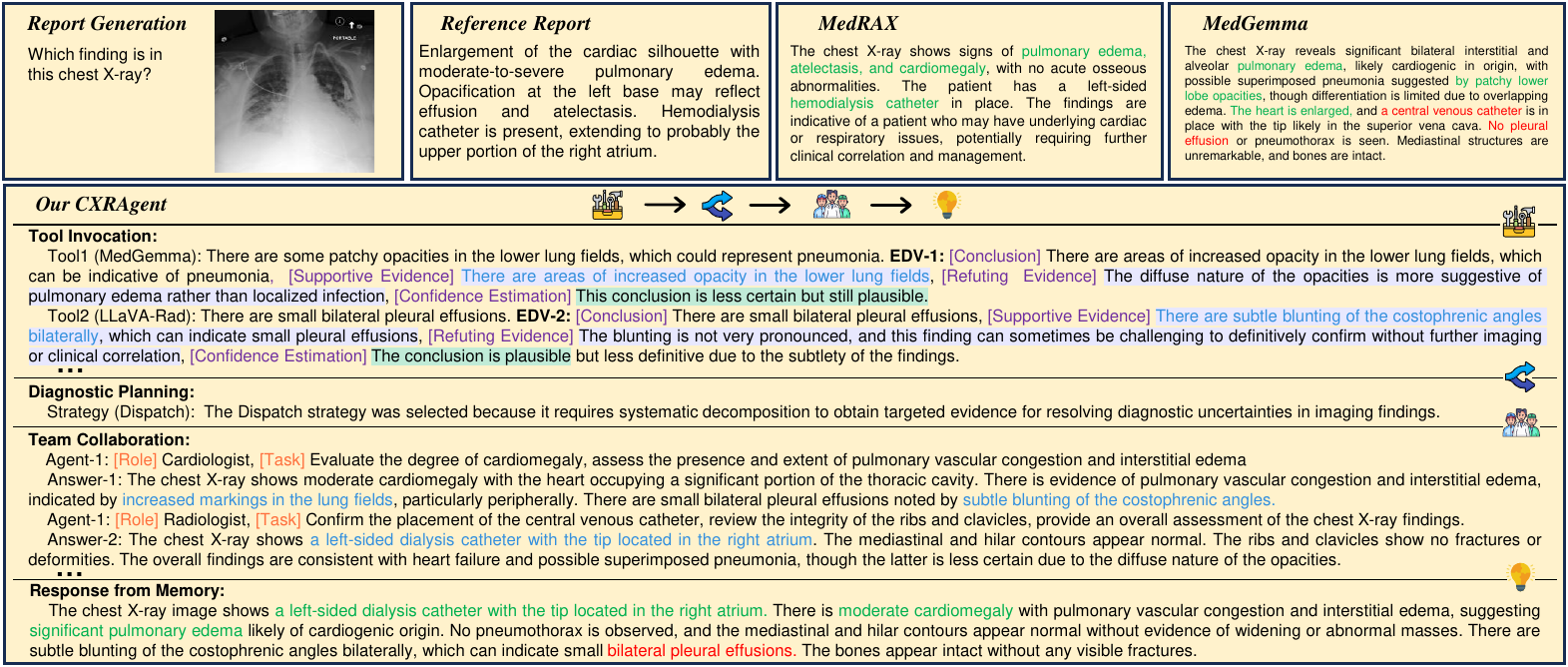}
\caption{\textcolor{black}{A  failure case study of CXRAgent versus state-of-the-art methods on report generation. \textbf{Top}: user query, reference report, and baseline predictions. \textbf{Bottom}: Prediction of CXRAgent. EDV sections are highlighted in purple with key evidence in blue and confidence highlighted in green. The roles and tasks of team members are in orange. Correct findings are in green and incorrect ones in red.}}
\label{failure}
\end{figure*}

\textcolor{black}{\textbf{Failure Case.} We present a failure case on Fig. \ref{failure}. As can be observed, the generated output exhibits notable inconsistencies with the ground truth: first, a spatial mislocalization of pleural effusion; and second, a failure to identify pulmonary opacification.  This performance degradation likely arises from the overlapping morphological characteristics of certain pathologies. For example, opacities can be indicative of either atelectasis or pleural effusion, leading to potential model confusion during visual feature interpretation. }


\subsection{Human Evaluation}
\textcolor{black}{We conducted a blinded pairwise comparison study involving five certified radiologists over 100 randomly selected samples. Our CXRAgent was compared against four state-of-the-art baselines: \textit{MedRAX}, \textit{CheXagent}, \textit{MedGemma}, and \textit{LLaVA-Rad}.  In each case, experts were presented with the original image, the reference report, and a randomized pair of generated reports, one from CXRAgent and one from a randomly selected baseline, while remaining blinded to the model identities.  The evaluation was conducted based on three key dimensions: 1) Completeness (comprehensiveness of anatomical coverage and findings); 2) Accuracy (factual correctness of identified findings and descriptions); and 3) Clinical Utility (the practical value in assisting the diagnostic decision-making such as reducing misdiagnoses and missed diagnoses, or improving efficiency). The win rates of CXRAgent against each method are shown in Table~\ref{tab:expert_eval}.    CXRAgent consistently outperforms other   models, achieving high average win rates for Completeness, Accuracy, and  Clinical Usability.  The most notable result is the  average win rate in Clinical Usability. This metric reflects the practical utility of the reports for medical decision-making. This superiority is primarily attributed to CXRAgent's capability to provide grounded visual evidence to justify its diagnostic findings.}

\begin{table}[h]
\centering
\small
\caption{\textcolor{black}{Expert evaluation results. The values are the average win rate and tie rate (\%) of CXRAgent against each method.}}
\label{tab:expert_eval_corrected}
\begin{tabular}{lcccccc}
\toprule
CXRAgent vs. & \multicolumn{2}{c}{{Completeness}} & \multicolumn{2}{c}{{Accuracy}} & \multicolumn{2}{c}{{Clinical Usability}} \\
\cmidrule(lr){2-3} \cmidrule(lr){4-5} \cmidrule(lr){6-7}
                      & {Win} & {Tie} & {Win} & {Tie} & {Win} & {Tie} \\
\midrule
CheXagent             & 58.6         & 4.0          & 52.8         & 1.2          & 54.4         & 2.4          \\
MedGemma              & 76.2         & 6.4          & 80.8         & 1.8          & 87.0         & 0.4          \\
MedRAX                & 85.6         & 5.2          & 84.2         & 2.8          & 85.2         & 1.8          \\
LLaVA-Rad             & 42.6         & 11.6         & 50.8         & 6.4          & 51.6         & 4.4          \\
\midrule
{Average}       & {65.8} & {6.8}  & {67.2} & {3.1}  & {69.6} & {2.3}  \\
\bottomrule
\label{tab:expert_eval}
\end{tabular}
\end{table}

\subsection{Discussion}
\textcolor{black}{Regarding the questions raised in the ``Introduction", we discuss how our study answers these questions:
1) Generalization across diverse CXR tasks: CXRAgent unifies task-specific and foundation model capabilities within a multi-stage agent framework. Experimental results on three benchmarks show that CXRAgent achieves robust generalization across disease identification, visual question answering, report generation and other CXR interpretation tasks. 2) Assessing and reconciling conflicting tool outputs:  The EDV evaluates each tool's output using supporting or refuting visual evidence, assigning confidence scores to ensure reliable downstream reasoning. This mechanism enables CXRAgent to effectively resolve conflicts among multiple tools, improving the accuracy and credibility of its diagnostic decisions. 3) Dynamic, team-based collaborative reasoning: Inspired by clinical MDT, CXRAgent dynamically assembles expert teams and coordinates their interactions according to task complexity and intermediate findings. This  collaboration performs context-aware planning and reasoning. Its effectiveness is validated in complex scenarios, achieving a RaTEScore of 0.569 for CXR report generation. By decomposing complex CXR interpretation into multi-stages, incorporating intermediate validation through EDV, and enabling collaborative reasoning, the system reduces error propagation, enhances robustness under tool conflicts, and generates structured, interpretable outputs. While this multi-stage framework introduces additional computational overhead compared with single-loop ReAct-style agents, it prioritizes diagnostic accuracy and interpretability (critical for clinical deployment). Moreover, the \textbf{[Skip]} strategy allows the system to bypass intensive collaboration in straightforward cases, maintaining computational adaptivity.} 
\textcolor{black}{
While CXRAgent shows strong potential to improve diagnostic accuracy and reliability, there are challenges for safe and effective real-time clinical deployment: }
\begin{itemize}
    \item \textcolor{black}{\textbf{Computational Overhead:} The multi-stage orchestration, iterative tool reasoning, and Evidence-driven Validator (EDV) introduce additional computational steps and latency compared with simpler single-loop agents. In high-throughput clinical environments, this overhead may limit real-time responsiveness, necessitating optimization strategies like model quantization.}

    \item \textcolor{black}{\textbf{Resource Requirements:} CXRAgent relies on a large multimodal LLM as the Director and multiple specialized diagnostic tools, requiring substantial GPU memory, high-speed processing. Such hardware requirements may not be available in hospitals, particularly in low-resource or rural environments, and may impose cost and infrastructure barriers for deployment.}
    
    \item \textcolor{black}{\textbf{Regulatory and Safety:} AI-assisted diagnostic systems must comply with strict regulatory standards, including traceability, auditability, and fail-safe mechanisms. Also, integration into existing clinical workflows requires maintaining physician oversight and enabling timely intervention in ambiguous cases. Protecting patient privacy and ensuring system stability are also critical for adoption.}
\end{itemize}

\section{Conclusion}
We introduced CXRAgent that conducts evidence-backed and  adaptive CXR interpretation through tool invocation, multi-stage reasoning, evidence-driven validation, and team-based collaboration. The core stages are: 1)  Tool Invocation: invoking diverse CXR-analysis tools with outputs standardized and verified by an evidence-driven validator; 2) Diagnostic Planning: flexibly planning  and assembling specialized expert teams based on task complexity and prior findings; and 3) Collaborative Decision-making:  consolidating expert insights and contextual memory into reliable, evidence-backed diagnostic conclusions. Extensive experiments across multiple CXR interpretation tasks demonstrate the superior reliability and  adaptability of CXRAgent  in handling  heterogeneous tool outputs and complex clinical scenarios, highlighting  its potential for real-world clinical deployment. 
\bibliographystyle{IEEEtran}
\bibliography{tmi}

@IEEEtranBSTCTL{IEEEexample:BSTcontrol,
  CTLmax_names_forced_etal = "6",
  CTLnames_show_etal = "1",
  CTLuse_article_number = "yes",
  CTLuse_paper = "yes",
  CTLuse_forced_etal = "yes"
}

@inproceedings{2024-green,
    title = "{GREEN}: Generative Radiology Report Evaluation and Error Notation",
    author = "Ostmeier, Sophie  and
      Xu, Justin  and
      Chen, Zhihong  and
      Varma, Maya  and
      Blankemeier, Louis  and
      Bluethgen, Christian  and
      Michalson, Arne Edward  and
      Moseley, Michael  and
      Langlotz, Curtis  and
      Chaudhari, Akshay S  and
      Delbrouck, Jean-Benoit",
    booktitle = "Findings of the Association for Computational Linguistics: EMNLP 2024",
    month = nov,
    year = "2024",
    pages = "374--390",
}

@inproceedings{radeval,
    title = "{R}ad{E}val: A framework for radiology text evaluation",
    author = "Xu, Justin  and
      Zhang, Xi  and
      Abderezaei, Javid  and
      Bauml, Julie  and
      Boodoo, Roger  and
      Haghighi, Fatemeh  and
      Ganjizadeh, Ali  and
      Brattain, Eric  and
      Van Veen, Dave  and
      Meng, Zaiqiao  and
      Eyre, David W  and
      Delbrouck, Jean-Benoit",
    booktitle = "Proceedings of the 2025 Conference on Empirical Methods in Natural Language Processing: System Demonstrations",
    month = nov,
    year = "2025",
    pages = "546--557",
}

@inproceedings{delbrouck-etal-2022-improving,
    title = "Improving the Factual Correctness of Radiology Report Generation with Semantic Rewards",
    author = "Delbrouck, Jean-Benoit  and
      Chambon, Pierre  and
      Bluethgen, Christian  and
      Tsai, Emily  and
      Almusa, Omar  and
      Langlotz, Curtis",
    booktitle = "Findings of the Association for Computational Linguistics: EMNLP 2022",
    month = dec,
    year = "2022",
    pages = "4348--4360",
}

@article{Zhao2026DeepRare,
  author    = {Zhao, Weike and Wu, Chaoyi and Fan, Yanjie and Qiu, Pengcheng and 
               Zhang, Xiaoman and Sun, Yuze and Zhou, Xiao and Zhang, Shuju and 
               Peng, Yu and Wang, Yanfeng and Sun, Xin and Zhang, Ya and 
               Yu, Yongguo and Sun, Kun and Xie, Weidi},
  title     = {An agentic system for rare disease diagnosis with traceable reasoning},
  journal   = {Nature},
  volume    = {651},
  pages     = {775--784},
  year      = {2026},
}

@INPROCEEDINGS{10145699,
  author={Nimalsiri, Wimukthi and Hennayake, Mahela and Rathnayake, Kasun and Ambegoda, Thanuja D. and Meedeniya, Dulani},
  booktitle={2023 3rd International Conference on Advanced Research in Computing}, 
  title={Automated Radiology Report Generation Using Transformers}, 
  year={2023},
  volume={},
  number={},
  pages={90-95},
  doi={10.1109/ICARC57651.2023.10145699}}

@INPROCEEDINGS{9682248,
  author={Kolonne, Shammi and Fernando, Chamodi and Kumarasinghe, Hashara and Meedeniya, Dulani},
  booktitle={2021 International Conference on Decision Aid Sciences and Application}, 
  title={MobileNetV2 Based Chest X-Rays Classification}, 
  year={2021},
  volume={},
  number={},
  pages={57-61},
  doi={10.1109/DASA53625.2021.9682248}}

@ARTICLE{11153468,
  author={Edirisinghe, Dasith and Nimalsiri, Wimukthi and Hennayake, Mahela and Meedeniya, Dulani and Lim, Gilbert},
  journal={IEEE Access}, 
  title={Chest X-Ray Report Generation Using Abnormality Guided Vision Language Model}, 
  year={2025},
  volume={13},
  number={},
  pages={157651-157673},
  }

@article{zheng2025end,
  title={End-to-End Agentic {RAG} System Training for Traceable Diagnostic Reasoning},
  author={Zheng, Qiaoyu and Sun, Yuze and Wu, Chaoyi and Zhao, Weike and Qiu, Pengcheng and Yu, Yongguo and Sun, Kun and Wang, Yanfeng and Zhang, Ya and Xie, Weidi},
  journal={arXiv preprint arXiv:2508.15746},
  year={2025}
}

@ARTICLE{peng2024integration,
  author={Peng, Qi and Cai, Yi and Liu, Jiankun and Zou, Quan and Chen, Xing and Zhong, Zheng and Wang, Zefeng and Xie, Jiayuan and Li, Qing},
  journal={IEEE Transactions on Medical Imaging}, 
  title={Integration of Multi-Source Medical Data for Medical Diagnosis Question Answering}, 
  year={2025},
  volume={44},
  number={3},
  pages={1373-1385},
 }

@ARTICLE{xie2025prompt,
  author={Xie, Xingyu and Zhao, Wenjie and Nan, Mu and Zhang, Zheng and Wu, Yaping and Zheng, Hairong and Liang, Dong and Wang, Meiyun and Hu, Zhanli},
  journal={IEEE Transactions on Medical Imaging}, 
  title={Prompt-Agent-Driven Integration of Foundation Model Priors for Low-Count PET Reconstruction}, 
  year={2025},
  volume={44},
  number={10},
  pages={4073-4086},
 }

@article{zhang2025patho,
  title={{Patho-AgenticRAG}: Towards multimodal agentic retrieval-augmented generation for pathology vlms via reinforcement learning},
  author={Zhang, Wenchuan and Guo, Jingru and Zhang, Hengzhe and Zhang, Penghao and Chen, Jie and Zhang, Shuwan and Zhang, Zhang and Yi, Yuhao and Bu, Hong},
  journal={arXiv preprint arXiv:2508.02258},
  year={2025}
}

@inproceedings{zhou2025mam,
    title = "{MAM}: Modular Multi-Agent Framework for Multi-Modal Medical Diagnosis via Role-Specialized Collaboration",
    author = "Zhou, Yucheng  and
      Song, Lingran  and
      Shen, Jianbing",
    booktitle = "Findings of the Association for Computational Linguistics: ACL 2025",
    month = jul,
    year = "2025",
    pages = "25319--25333",
}

@article{DeepMedix-R1,
  title={A Foundation Model for Chest X-ray Interpretation with Grounded Reasoning via Online Reinforcement Learning},
  author={Lin, Qika and Zhu, Yifan and Pu, Bin and Huang, Ling and Luo, Haoran and Ma, Jingying and Peng, Zhen and Zhao, Tianzhe and Xu, Fangzhi and Zhang, Jian and He, Kai and Ou, Zhonghong and Mishra, Swapnil and Feng, Mengling},
  journal={arXiv preprint arXiv:2509.03906},
  year={2025}
}

@inproceedings{irvin2019chexpert,
  title={Chexpert: A large chest radiograph dataset with uncertainty labels and expert comparison},
  author={Irvin, Jeremy and Rajpurkar, Pranav and Ko, Michael and Yu, Yifan and Ciurea-Ilcus, Silviana and Chute, Chris and Marklund, Henrik and Haghgoo, Behzad and Ball, Robyn and Shpanskaya, Katie and others},
  booktitle={Proceedings of the AAAI conference on artificial intelligence},
  volume={33},
  number={01},
  pages={590--597},
  year={2019}
}

@article{medgemma-hf,
  title={MedGemma Technical Report},
  author={Sellergren, Andrew and Kazemzadeh, Sahar and Jaroensri, Tiam and Kiraly, Atilla and Traverse, Madeleine and Kohlberger, Timo and Xu, Shawn and Jamil, Fayaz and Hughes, Cían and Lau, Charles and Chen, Justin and Mahvar, Fereshteh and Yatziv, Liron and Chen, Tiffany and Sterling, Bram and Baby, Stefanie Anna and Baby, Susanna Maria and Lai, Jeremy and Schmidgall, Samuel and Yang, Lu and Chen, Kejia and Bjornsson, Per and Reddy, Shashir and Brush, Ryan and Philbrick, Kenneth and Asiedu, Mercy and Mezerreg, Ines and Hu, Howard and Yang, Howard and Tiwari, Richa and Jansen, Sunny and Singh, Preeti and Liu, Yun and Azizi, Shekoofeh and Kamath, Aishwarya and Ferret, Johan and Pathak, Shreya and Vieillard, Nino and Merhej, Ramona and Perrin, Sarah and Matejovicova, Tatiana and Ramé, Alexandre and Riviere, Morgane and Rouillard, Louis and Mesnard, Thomas and Cideron, Geoffrey and Grill, Jean-Bastien and Ramos, Sabela and Yvinec, Edouard and Casbon, Michelle and Buchatskaya, Elena and Alayrac, Jean-Baptiste and Lepikhin, Dmitry and Feinberg, Vlad and Borgeaud, Sebastian and Andreev, Alek and Hardin, Cassidy and Dadashi, Robert and Hussenot, Léonard and Joulin, Armand and Bachem, Olivier and Matias, Yossi and Chou, Katherine and Hassidim, Avinatan and Goel, Kavi and Farabet, Clement and Barral, Joelle and Warkentin, Tris and Shlens, Jonathon and Fleet, David and Cotruta, Victor and Sanseviero, Omar and Martins, Gus and Kirk, Phoebe and Rao, Anand and Shetty, Shravya and Steiner, David F and Kirmizibayrak, Can and Pilgrim, Rory and Golden, Daniel and Yang, Lin},
  journal={arXiv preprint arXiv:2507.05201},
  year={2025}
}

@article{Lingshu,
      title={Lingshu: A Generalist Foundation Model for Unified Multimodal Medical Understanding and Reasoning}, 
      author={LASA Team and Weiwen Xu and Hou Pong Chan and Long Li and Mahani Aljunied and Ruifeng Yuan and Jianyu Wang and Chenghao Xiao and Guizhen Chen and Chaoqun Liu and Zhaodonghui Li and Yu Sun and Junao Shen and Chaojun Wang and Jie Tan and Deli Zhao and Tingyang Xu and Hao Zhang and Yu Rong},
      year={2025},
      journal={arXiv preprint arXiv:2506.07044},
}

@inproceedings{yao2023react,
  title={{ReAct}: Synergizing reasoning and acting in language models},
  author={Yao, Shunyu and Zhao, Jeffrey and Yu, Dian and Du, Nan and Shafran, Izhak and Narasimhan, Karthik and Cao, Yuan},
  booktitle={International Conference on Learning Representations},
  year={2023}
}

@inproceedings{yue2024clinicalagent,
  title={{ClinicalAgent:} Clinical trial multi-agent system with large language model-based reasoning},
  author={Yue, Ling and Xing, Sixue and Chen, Jintai and Fu, Tianfan},
  booktitle={Proceedings of the 15th ACM International Conference on Bioinformatics, Computational Biology and Health Informatics},
  pages={1--10},
  year={2024}
}

@article{huang2025biomni,
  title={Biomni: A General-Purpose Biomedical {AI} Agent},
  author={Huang, Kexin and Zhang, Serena and Wang, Hanchen and Qu, Yuanhao and Lu, Yingzhou and Roohani, Yusuf and Li, Ryan and Qiu, Lin and Zhang, Junze and Di, Yin and others},
  journal={bioRxiv},
  pages={2025--05},
  year={2025},
}

@article{jin2024agentmd,
  author    = {Qiao Jin and Zhizheng Wang and Yifan Yang and Qingqing Zhu and Donald Wright and Thomas Huang and Nikhil Khandekar and Nicholas Wan and Xuguang Ai and W. John Wilbur and Zhe He and R. Andrew Taylor and Qingyu Chen and Zhiyong Lu},
  title     = {{AgentMD}: Empowering language agents for risk prediction with large-scale clinical tool learning},
  journal   = {Nature Communications},
  year      = {2025},
  volume    = {16},
  number    = {1},
  pages     = {9377},
}

@article{mao2025ct,
  title={{CT-Agent:} A Multimodal-LLM Agent for {3D CT} Radiology Question Answering},
  author={Mao, Yuren and Xu, Wenyi and Qin, Yuyang and Gao, Yunjun},
  journal = { Sci. China Inf. Sci.},
  volume  = {69},
  pages   = {150107},
  year    = {2026},
  month   = {Apr.},
}

@ARTICLE{Detection,
  author={Lian, Jie and Liu, Jingyu and Zhang, Shu and Gao, Kai and Liu, Xiaoqing and Zhang, Dingwen and Yu, Yizhou},
  journal={IEEE Transactions on Medical Imaging}, 
  title={A Structure-Aware Relation Network for Thoracic Diseases Detection and Segmentation}, 
  year={2021},
  volume={40},
  number={8},
  pages={2042-2052}}

@ARTICLE{Tokenmixer,
  author={Yang, Yan and Yu, Jun and Fu, Zhenqi and Zhang, Ke and Yu, Ting and Wang, Xianyun and Jiang, Hanliang and Lv, Junhui and Huang, Qingming and Han, Weidong},
  journal={IEEE Transactions on Medical Imaging}, 
  title={{Token-Mixer}: Bind Image and Text in One Embedding Space for Medical Image Reporting}, 
 year={2024},
  volume={43},
  number={11},
  pages={4017-4028},
 }

@InProceedings{24,
    author    = {Tanida, Tim and M\"uller, Philip and Kaissis, Georgios and Rueckert, Daniel},
    title     = {Interactive and Explainable Region-Guided Radiology Report Generation},
    booktitle = {Proceedings of the IEEE Conference on Computer Vision and Pattern Recognition},
    year      = {2023},
    pages     = {7433-7442}
}

@article{wu2023generalistfoundationmodelradiology,
  author    = {Wu, Chaoyi and Zhang, Xiaoman and Zhang, Ya and Hui, Hui and Wang, Yanfeng and Xie, Weidi},
  title     = {Towards generalist foundation model for radiology by leveraging web-scale {2D\&3D} medical data},
  journal   = {Nature Communications},
  volume    = {16},
  number    = {1},
  pages     = {7866},
  year      = {2025},
}

@article{ark,
  title={A fully open {AI} foundation model applied to chest radiography},
  author={Ma, DongAo and Pang, Jiaxuan and Gotway, Michael B and Liang, Jianming},
  journal={Nature},
  pages={1--11},
  year={2025},
}

@inproceedings{zhao2024ratescore,
  title={{RaTEScore}: A Metric for Radiology Report Generation},
  author={Zhao, Weike and Wu, Chaoyi and Zhang, Xiaoman and Zhang, Ya and Wang, Yanfeng and Xie, Weidi},
  booktitle={Proceedings of the 2024 Conference on Empirical Methods in Natural Language Processing},
  pages={15004--15019},
  year={2024}
}

@article{WANG2023,
title = {{R2GenGPT}: Radiology Report Generation with frozen LLMs},
journal = {Meta-Radiology},
author = {Zhanyu Wang and Lingqiao Liu and Lei Wang and Luping Zhou},
volume = {1},
number = {3},
pages = {100033},
year = {2023},
issn = {2950-1628},
}

@article{NICOLSON2024101585,
title = {Longitudinal data and a semantic similarity reward for chest X-ray report generation},
journal = {Informatics in Medicine Unlocked},
volume = {50},
pages = {101585},
year = {2024},
issn = {2352-9148},
author = {Aaron Nicolson and Jason Dowling and Douglas Anderson and Bevan Koopman},
}

@inproceedings{CXRMate-ED,
    title = "The Impact of Auxiliary Patient Data on Automated Chest {X}-Ray Report Generation and How to Incorporate It",
    author = "Nicolson, Aaron  and Zhuang, Shengyao and Dowling, Jason and Koopman, Bevan",
    booktitle = "Proceedings of the 63rd Annual Meeting of the Association for Computational Linguistics (Volume 1: Long Papers)",
    month = jul,
    year = "2025",
    pages = "177--203",
}

@article{cvt2distilgpt2,
title = {Improving chest X-ray report generation by leveraging warm starting},
journal = {Artificial Intelligence in Medicine},
volume = {144},
pages = {102633},
year = {2023},
issn = {0933-3657},
author = {Aaron Nicolson and Jason Dowling and Bevan Koopman},
}

@inproceedings{libra,
    title = "Libra: Leveraging Temporal Images for Biomedical Radiology Analysis",
    author = "Zhang, Xi  and
      Meng, Zaiqiao  and
      Lever, Jake  and
      Ho, Edmond S. L.",
    booktitle = "Findings of the Association for Computational Linguistics: ACL 2025",
    month = jul,
    year = "2025",
    pages = "17275--17303",
}

@article{Janus-Pro,
  title={Janus-Pro: Unified Multimodal Understanding and Generation with Data and Model Scaling},
  author={Chen, Xiaokang and Wu, Zhiyu and Liu, Xingchao and Pan, Zizheng and Liu, Wen and Xie, Zhenda and Yu, Xingkai and Ruan, Chong},
  journal={arXiv preprint arXiv:2501.17811},
  year={2025}
}

@article{MedVersa,
author = {Hong-Yu Zhou  and Julián Nicolás Acosta  and Subathra Adithan  and Suvrankar Datta  and Eric J. Topol  and Pranav Rajpurkar },
title = {MedVersa: A Generalist Foundation Model for Diverse Medical Imaging Tasks},
journal = {NEJM AI},
pages = {AIoa2500595},
year = {2026},
}

@inproceedings{ifcc,
    title = "Improving Factual Completeness and Consistency of Image-to-Text Radiology Report Generation",
    author = "Miura, Yasuhide  and
      Zhang, Yuhao  and
      Tsai, Emily  and
      Langlotz, Curtis  and
      Jurafsky, Dan",
    booktitle = "Proceedings of the 2021 Conference of the North American Chapter of the Association for Computational Linguistics: Human Language Technologies",
    month = jun,
    year = "2021",
    pages = "5288--5304",
}

@ARTICLE{liang2025medfilip,
  author={Liang, Xinjie and Li, Xiangyu and Li, Fanding and Jiang, Jie and Dong, Qing and Wang, Wei and Wang, Kuanquan and Dong, Suyu and Luo, Gongning and Li, Shuo},
  journal={IEEE Journal of Biomedical and Health Informatics}, 
  title={MedFILIP: Medical Fine-Grained Language-Image Pre-Training}, 
  year={2025},
  volume={29},
  number={5},
  pages={3587-3597},
}

@inproceedings{moor2023med,
  title={Med-flamingo: a multimodal medical few-shot learner},
  author={Moor, Michael and Huang, Qian and Wu, Shirley and Yasunaga, Michihiro and Dalmia, Yash and Leskovec, Jure and Zakka, Cyril and Reis, Eduardo Pontes and Rajpurkar, Pranav},
  booktitle={Machine Learning for Health (ML4H)},
  pages={353--367},
  year={2023},
  organization={PMLR}
}

@ARTICLE{MVQA,
  author={Yu, Ting and Ge, Binhui and Wang, Shuhui and Yang, Yan and Huang, Qingming and Yu, Jun},
  journal={IEEE Journal of Biomedical and Health Informatics}, 
  title={Consistency Conditioned Memory Augmented Dynamic Diagnosis Model for Medical Visual Question Answering}, 
  year={2025},
  volume={29},
  number={2},
  pages={1357-1370},}

@inproceedings{fallahpour2025medrax,
  title={{MedRAX:} Medical Reasoning Agent for Chest X-ray},
  author={Fallahpour, Adibvafa and Ma, Jun and Munim, Alif and Lyu, Hongwei and Wang, Bo},
  booktitle={Proc. 42nd Int. Conf. Mach. Learn. (ICML)},
  pages = 	 {15661--15676},
  year = 	 {2025},
  volume = 	 {267},
}

@inproceedings{
kim2024mdagents,
title={{MDA}gents: An Adaptive Collaboration of {LLM}s for Medical Decision-Making},
author={Yubin Kim and Chanwoo Park and Hyewon Jeong and Yik Siu Chan and Xuhai Xu and Daniel McDuff and Hyeonhoon Lee and Marzyeh Ghassemi and Cynthia Breazeal and Hae Won Park},
booktitle={The Thirty-eighth Annual Conference on Neural Information Processing Systems},
year={2024},
}

@inproceedings{li2024mmedagent,
    title = "{MM}ed{A}gent: Learning to Use Medical Tools with Multi-modal Agent",
    author = "Li, Binxu  and
      Yan, Tiankai  and
      Pan, Yuanting  and
      Luo, Jie  and
      Ji, Ruiyang  and
      Ding, Jiayuan  and
      Xu, Zhe  and
      Liu, Shilong  and
      Dong, Haoyu  and
      Lin, Zihao  and
      Wang, Yixin",
    booktitle = "Findings of the Association for Computational Linguistics: EMNLP 2024",
    month = nov,
    year = "2024",
    pages = "8745--8760",
}

@article{bannur2024maira,
  title={{MAIRA-2:} Grounded radiology report generation},
  author={Bannur, Shruthi and Bouzid, Kenza and Castro, Daniel C and Schwaighofer, Anton and Thieme, Anja and Bond-Taylor, Sam and Ilse, Maximilian and P{\'e}rez-Garc{\'\i}a, Fernando and Salvatelli, Valentina and Sharma, Harshita and others},
  journal={arXiv preprint arXiv:2406.04449},
  year={2024}
}

@inproceedings{cohen2022torchxrayvision,
  title={TorchXRayVision: A library of chest X-ray datasets and models},
  author={Cohen, Joseph Paul and Viviano, Joseph D and Bertin, Paul and Morrison, Paul and Torabian, Parsa and Guarrera, Matteo and Lungren, Matthew P and Chaudhari, Akshay and Brooks, Rupert and Hashir, Mohammad and others},
  booktitle={International Conference on Medical Imaging with Deep Learning},
  pages={231--249},
  year={2022},
}

@article{ma2024segment,
  title={Segment anything in medical images},
  author={Ma, Jun and He, Yuting and Li, Feifei and Han, Lin and You, Chenyu and Wang, Bo},
  journal={Nature Communications},
  volume={15},
  number={1},
  pages={654},
  year={2024},
}

@inproceedings{
tang2024medagents,
title={{MedAgents}: Large Language Models as Collaborators for Zero-shot Medical Reasoning},
author={Xiangru Tang and Anni Zou and Zhuosheng Zhang and Ziming Li and Yilun Zhao and Xingyao Zhang and Arman Cohan and Mark Gerstein},
booktitle={ICLR 2024 Workshop on Large Language Model (LLM) Agents},
year={2024},
}

@inproceedings{
chen2024chexagent,
title={{CheXagent}: Towards a Foundation Model for Chest X-Ray Interpretation},
author={Zhihong Chen and Maya Varma and Jean-Benoit Delbrouck and Magdalini Paschali and Louis Blankemeier and Dave Van Veen and Jeya Maria Jose Valanarasu and Alaa Youssef and Joseph Paul Cohen and Eduardo Pontes Reis and Emily Tsai and Andrew Johnston and Cameron Olsen and Tanishq Mathew Abraham and Sergios Gatidis and Akshay S Chaudhari and Curtis Langlotz},
booktitle={AAAI 2024 Spring Symposium on Clinical Foundation Models},
year={2024},
}

@article{DoctorAgent,
  title={{DoctorAgent-RL: A } Multi-Agent Collaborative Reinforcement Learning System for Multi-Turn Clinical Dialogue},
  author={Feng, Yichun and Wang, Jiawei and Zhou, Lu and Li, Yixue},
  journal={arXiv preprint arXiv:2505.19630},
  year={2025}
}

@article{MIMIC,
  author       = {Alistair E. W. Johnson and
                  Tom J. Pollard and
                  Seth J. Berkowitz and
                  Nathaniel R. Greenbaum and
                  Matthew P. Lungren and
                  Chih{-}ying Deng and
                  Roger G. Mark and
                  Steven Horng},
  title        = {{MIMIC-CXR:} {A} large publicly available database of labeled chest radiographs},
  journal      = {Scientific Data},
  volume       = {6},
  year         = {2019},
  pages        = {317},
}

@inproceedings{li2023llava,
  title={{LLaVA-Med}: Training a large language-and-vision assistant for biomedicine in one day},
  author={Li, Chunyuan and Wong, Cliff and Zhang, Sheng and Usuyama, Naoto and Liu, Haotian and Yang, Jianwei and Naumann, Tristan and Poon, Hoifung and Gao, Jianfeng},
  booktitle={Advances in Neural Information Processing Systems},
  volume={36},
  pages={28541--28564},
  year={2023}
}

@article{zambrano2025clinically,
  title={A clinically accessible small multimodal radiology model and evaluation metric for chest X-ray findings},
  author={Zambrano Chaves and Juan Manuel and Huang, Shih-Cheng and Xu, Yanbo and Xu, Hanwen and Usuyama, Naoto and Zhang, Sheng and Wang, Fei and Xie, Yujia and Khademi, Mahmoud and Yang, Ziyi and others},
  journal={Nature Communications},
  volume={16},
  number={1},
  pages={3108},
  year={2025},
}

@article{HU2024103279,
title = {Interpretable medical image Visual Question Answering via multi-modal relationship graph learning},
journal = {Medical Image Analysis},
volume = {97},
pages = {103279},
year = {2024},
author = {Xinyue Hu and Lin Gu and Kazuma Kobayashi and Liangchen Liu and Mengliang Zhang and Tatsuya Harada and Ronald M. Summers and Yingying Zhu},
}

@InProceedings{pan2025medvlm,
        author = { Pan, Jiazhen AND Liu, Che AND Wu, Junde AND Liu, Fenglin AND Zhu, Jiayuan AND Li, Hongwei Bran AND Chen, Chen AND Ouyang, Cheng AND Rueckert, Daniel},
        title = {{MedVLM-R1: Incentivizing Medical Reasoning Capability of Vision-Language Models {(VLMs)} via Reinforcement Learning}},
        booktitle = {proceedings of Medical Image Computing and Computer Assisted Intervention -- MICCAI 2025},
        year = {2025},
        volume = {LNCS 15966},
        month = {September},
        page = {337 -- 347}
}

@article{PIXEL,
  author    = {Kaiming Dong and Yuxiao Cheng and Kunlun He and Jinli Suo},
  title     = {A generative model uses healthy and diseased image pairs for pixel-level chest X-ray pathology localization},
  journal   = {Nature Biomedical Engineering},
  volume = {10},
  pages = {325--337},
  year = {2026},
}

@inproceedings{FAVP,
  author       = {Ting Yu and
                  Zixuan Tong and
                  Jun Yu and
                  Ke Zhang},
  title        = {Fine-grained Adaptive Visual Prompt for Generative Medical Visual Question Answering},
  booktitle    = {the AAAI Conference on Artificial Intelligence},
  pages        = {9662--9670},
  year         = {2025},
}

@inproceedings{COD,
    title = "{C}o{D}, Towards an Interpretable Medical Agent using Chain of Diagnosis",
    author = "Chen, Junying  and
      Gui, Chi  and
      Gao, Anningzhe  and
      Ji, Ke  and
      Wang, Xidong  and
      Wan, Xiang  and
      Wang, Benyou",
    booktitle = "Findings of the Association for Computational Linguistics: ACL 2025",
    month = jul,
    year = "2025",
    pages = "14345--14368",
}
\end{document}